\pdfoutput=1

\documentclass[11pt]{article}

\usepackage{acl}

\usepackage{times}
\usepackage{latexsym}

\usepackage{amssymb}
\usepackage{graphicx}
\usepackage{subcaption}
\usepackage{array}
\usepackage{multirow}
\usepackage{booktabs}

\usepackage{enumitem}

\usepackage[T1]{fontenc}

\usepackage[utf8]{inputenc}

\usepackage{microtype}

\usepackage{amsmath}

\title{Dialogue Term Extraction using Transfer Learning and \\ Topological Data Analysis}

\author{Renato Vukovic, 
{\bf Michael Heck}, 
{\bf Benjamin Ruppik} \\
{\bf Carel van Niekerk}, 
{\bf Marcus Zibrowius}, 
{\bf Milica Ga\v{s}i\'{c} } \\
  Heinrich Heine University Düsseldorf, Germany \\
  \texttt{\{renato.vukovic, heckmi, ruppik, niekerk, marcus.zibrowius, gasic\}@hhu.de}
}
  
\DeclareMathOperator{\VR}{VR} 

\newcommand{\MASK}{\textrm{MASK}}

\newcommand{\norm}[1]{\left\lVert#1\right\rVert} 

\usepackage{tikz}
\newcommand\mybox[2][]{\tikz[overlay]\node[fill=blue!20,inner sep=1pt, anchor=text, rectangle, rounded corners=1mm,#1] {#2};\phantom{#2}}

\begin{document}
\maketitle
\begin{abstract}
Goal oriented dialogue systems were originally designed as a natural language interface to a fixed data-set of entities that users might inquire about, further described by domain, slots and values.
As we move towards adaptable dialogue systems where knowledge about domains, slots and values may change, there is an increasing need to automatically extract these terms from raw dialogues or related non-dialogue data on a large scale. In this paper, we take an important step in this direction by exploring different features that can enable systems to discover realizations of domains, slots and values in dialogues in a purely data-driven fashion. 
The features that we examine stem from word embeddings, language modelling features, as well as topological features of the word embedding space.
To examine the utility of each feature set, we train a seed model based on the widely used MultiWOZ data-set.
Then, we apply this model to a different corpus, the Schema-Guided Dialogue data-set. Our method outperforms the previously proposed approach that relies solely on word embeddings. We also demonstrate that each of the features is responsible for discovering different kinds of content.
We believe our results warrant further research towards ontology induction, and continued harnessing of topological data analysis for dialogue and natural language processing research.
\end{abstract}

\section{Introduction}
\label{sec:introduction}

Dialogue systems are becoming increasingly popular as natural language interfaces to complex services. 
Goal-oriented dialogue systems, which we see as the main area of application of the results presented here, are intended to be capable of conversing with a user to solve one or more tasks.
They need to provide factual information and plan ahead over the course of multiple turns of dialogue.
Thus, they differ fundamentally from chat-based dialogue systems, which aim to engage the user in interesting conversation by offering entertainment.
Chat-based systems have been successfully trained using fully end-to-end approaches founded on large pretrained models~\cite{adiwardana2020meena, Lin2020caire, zhang-etal-2020-dialogpt, thoppilan2022lamda}.
In contrast, state-of-the-art goal-oriented dialogue systems continue to rely on a 
pre-defined \textit{ontology}: a database comprising domains (i.e., general topics for interaction), slots (constructs belonging to a particular topic), and values (concrete instantiations of such constructs) \cite{ultes-etal-2017-pydial, zhu-etal-2020-convlab, kulhanek-etal-2021-augpt, peng-etal-2021-soloist, lee-2021-mttod, he2022galaxy}.

Consequently, state-of-the-art goal-oriented dialogue systems still have a high reliance on manual labour. 
Firstly, the underlying ontology needs to be manually designed for each domain of conversation~\cite{milward2003ontology}. 
Secondly, the dialogue system needs to learn from a certain amount of dialogue data labelled with concepts from that ontology in order to recognize and understand these concepts in context~\cite{young2013pomdp}.
This manual annotation is again challenging, time-consuming and expensive~\cite{budzianowski2018multiwoz}.
There is thus a strong need for methods that can automate \textit{ontology construction} from raw data.
Moreover, ontology construction from raw dialogue data would have two-fold benefits: the dialogue data would be labelled automatically as the ontology is constructed, thus rendering any human involvement unnecessary.

In this work, we concentrate exclusively on the first step of ontology construction: \textit{term extraction}. The terms relate to regions of importance in the raw text. 
The subsequent steps of ontology construction, which we do not consider here, usually involve some form of clustering to boil down the extracted terms to a smaller number of concepts before they are finally organized into a full ontology.

Traditionally, term extraction begins by extracting terms based on frequency, in a way that aims to maximize recall~\cite{nakagawa2002simple,wermter2006you}. As frequency alone is a fairly primitive feature, this first step has close to zero precision and typically results in far too many terms. 
This makes further substantial filtering necessary within the term extraction step~\cite{frantzi1999c}. 
Filtering typically relies on heuristics or pre-existing natural language processing (NLP) models that have been trained on unrelated data, e.g., semantic parsers~\cite{bourigault1999term,aubin2006improving}. 
Heuristics as well as NLP models require substantial amounts of linguistic expertise to be created. 
%

In this work, we take a purely data-driven approach toward dialogue term detection to circumvent these limitations. 
The high dimensional data spaces arising from word embeddings are hard to understand and visualize.
Topological data analysis (TDA) is a collection of mathematical tools which provides measurements of the geometry of high-dimensional point clouds at various scales.
The major advantage of topological features is their invariance under small deformations and rotations, as opposed to the coordinates of the embedding vectors.
This leads to characteristics that are very generalizable and not dependent on the exact data set used for training.
The utility of TDA for NLP and dialogue modelling in particular are still under-explored.
We believe that information that can be gathered using topological methods has considerable predictive power concerning term extraction, which to the best of our knowledge we exploit with this work for the very first time.

Starting from the approach of~\citet{qiu-structure-2022}, we train a BIO-tagging~\cite{ramshaw-marcus-1995-text} model on the widely used MultiWOZ~\cite{budzianowski2018multiwoz} data-set as the seed set by fine-tuning general purpose large pre-trained language models. Our BIO-tagger accepts various features as input, all of which uniquely contribute to solving the task. 
We measure the zero-shot transfer ability of our proposed models on the Schema-Guided Dialogue~\cite{rastogi2020towards} data-set, another well-established large-scale corpus for dialogue modelling. 
Our contributions are as follows:
\begin{itemize}[leftmargin=*]
    \setlength\itemsep{-0.2em}
    \item We present novel features to solve the term extraction task. Our experimental results show significant improvements over a strong baseline, a recently proposed model that only takes contextual word embeddings as input.
    \item We demonstrate the suitability of masked language modelling scores to predict relevant terms.
    \item We exhibit the suitability of a range of topological features of neighbourhoods of word vectors to predict terms of relevance, including terms that are not present in the original seed training set.
    \item We make our code publicly available.\footnote{\url{http://doi.org/10.5281/zenodo.6858565}}
\end{itemize}

Our proposed method for term extraction leverages semantics as well as information gained from topological data analysis.
No element of our approach requires linguistic knowledge, nor do we rely on any heuristics.
Our models are either trained from scratch using a seed data-set, or leverage the predictive power of pre-trained and then fine-tuned large general purpose language models.
These models learn via self-supervision on large corpora, and our additional training only requires a moderate amount of labelled seed data.


\section{Related Work}
\label{sec:related}

It is normally assumed that the \emph{ontology} is provided and built independently of the dialogue system. For instance, in information seeking dialogue systems, this would be a structured representation of the database.  Approaches to ontology learning from texts generally involve enriching a small ontology with new concepts and new relationships using text mining methods such as linguistic techniques and lexico-syntactic patterns~\cite{pape06, agam08}, clustering techniques~\cite{aahm00, wits05}, statistical techniques~\cite{skfi03} and association rules~\cite{boab05, gubk05}. The majority of these methods require some form of human intervention.
The potential of machine learning in this area has been demonstrated in the Never-Ending Language Learning~(NELL) project~\cite{mcht15}. NELL learns factual knowledge from years of self-supervised experience in harvesting the web, using previously learned knowledge to improve subsequent learning.

In the pipeline of knowledge base construction, term extraction is typically the first step.
One example of a term extractor is presented in~\cite{sclano-termextractor-2007}.
It uses a part-of-speech (POS) tagger to select nouns, verbs and adjectives to which a number of heuristic frequency-based probabilistic models are applied to select term candidates. WordNet~\cite{WordNet:Hardcopy} is employed to handle misspellings.
A number of more recent methods for knowledge base construction start with a similar approach as~\citet{sclano-termextractor-2007}. In~\cite{romero-quasimodo-2020} we can also see heavy reliance on frequency, the use of dependency parsers in~\cite{nguyen-advanced-2021}, as well as rules based on lexical and numerical features and the use of WordNet as in~\cite{chu-tifi-2019}. 

A notable example of dialogue ontology induction is presented in~\cite{hudecek-etal-2021-discovering}, where a rule-based semantic parser is used as a starting point to propose an initial set of concepts. 
A more data-driven approach is presented by~\citet{qiu-structure-2022} who proposed training a BIO-tagger on fine-tuned contextual embeddings to induce slots. 
The approach is validated on MultiWOZ via leave-one-out domain experiments. We take this work as a starting point. 
In very recent work, \citet{yu-unsupervised-schema-2022} propose ontology induction using language modelling attention maps and regularized probabilistic context free grammar to detect regions of interest in text, followed by clustering. This work is complementary to ours, and it would be interesting to explore 
its combination with our proposal.
The `Beyond domain APIs' track of the 9th dialog system technology challenge (DSTC9)~\cite{gunasekara2020overview} aimed to remove friction in task-oriented dialogue systems where users might issue a request that is out of a system's scope. 
While DSTC9 aimed to integrate non-dialogue data into dialogue, none of the challenge submissions attempted ontology construction or expansion.

Topological data analysis remains largely underutilized in natural language processing.
One notable exception is the work presented by~\citet{jakubowski2020topology}. 
It shows that the Wasserstein norm of degree zero persistence of punctured neighbourhoods in a static word embedding correlates with the polysemy of a word. 
\citet{fraudfromabstractsTDA} apply persistent homology to word embedding point clouds with the goal of distinguishing fraudulent from genuine scientific publications.
Their best performing model utilizes persistence features derived from time-delay embeddings of term frequency data.
\citet{kushnareva-etal-2021-artificial} compute persistent homology of a filtered graph constructed from the attention maps of a pre-trained language model and harness the features for an artificial text detection task. 

\begin{figure}
    \centering
    \includegraphics[width=0.8\linewidth]{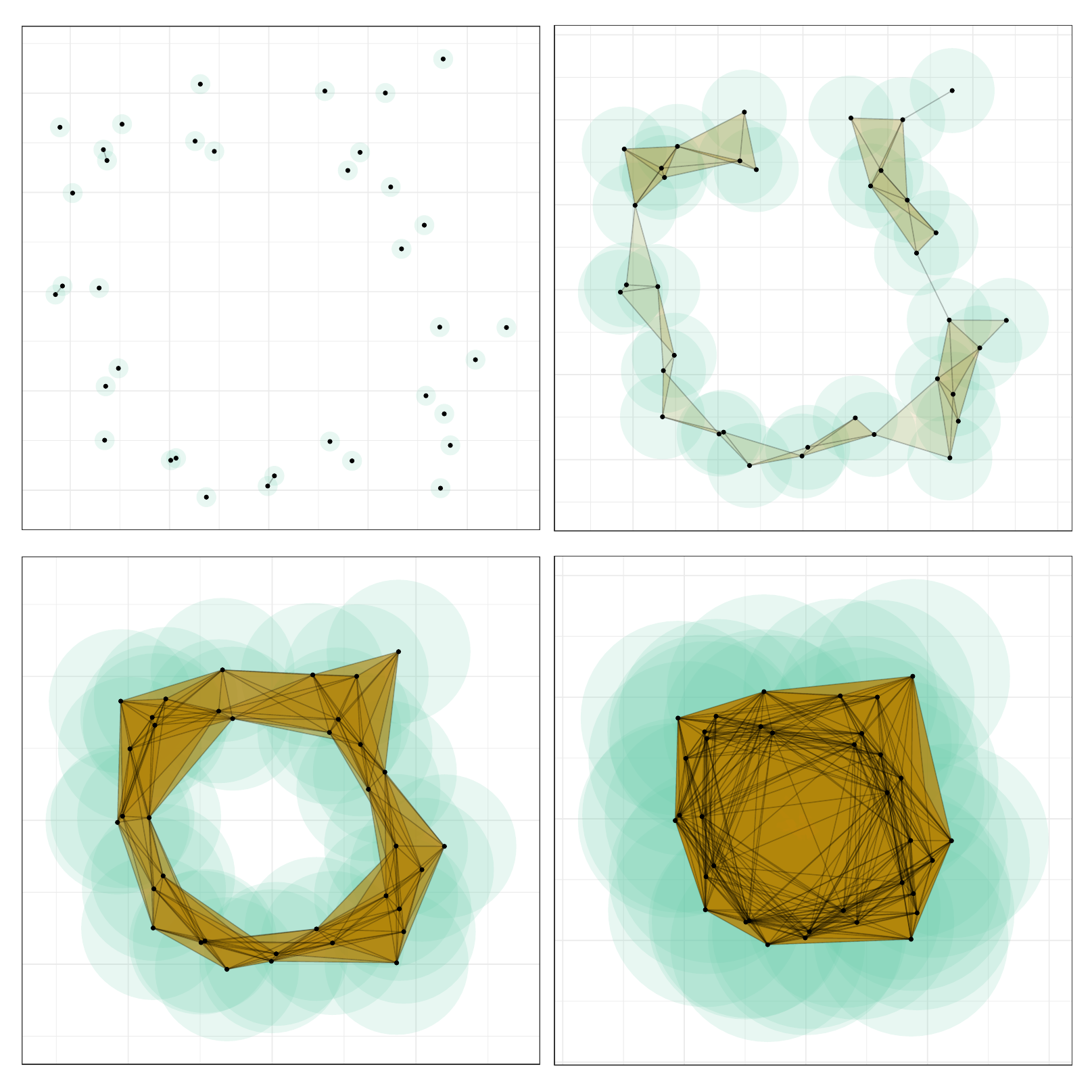}
    \caption{
        Illustration of the Vietoris-Rips complex $\VR_{\varepsilon}$ for four different values of \(\varepsilon\).
    }
    \label{fig:vietoris_rips_complex}
\end{figure}

\begin{figure*}
    \centering
    \begin{subfigure}[b]{0.3\textwidth}
        \centering
        \includegraphics[page=1, trim=24.0cm 0.0cm 2.0cm 0.0cm, clip=true, width=1.00\linewidth,]{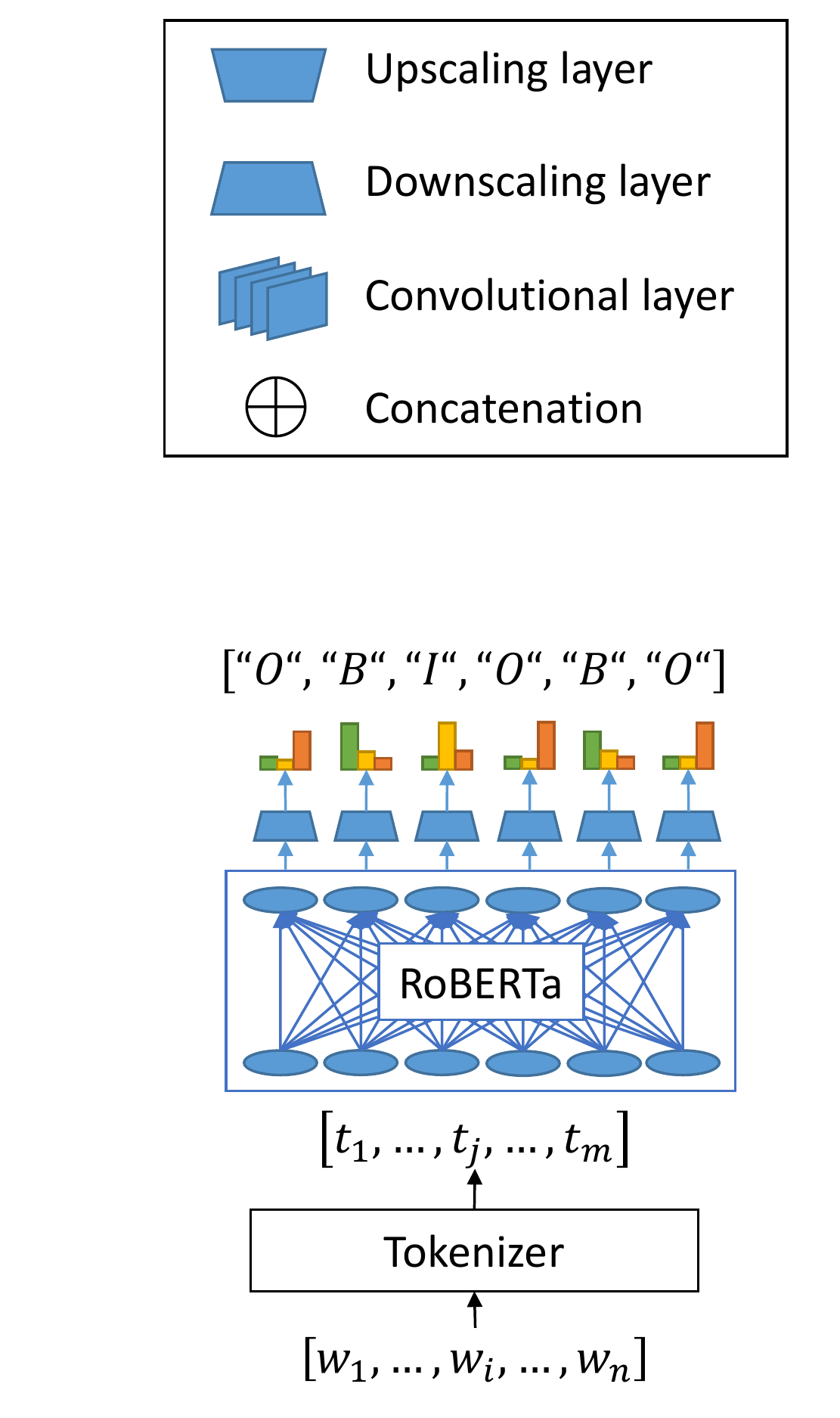}
        \caption{Baseline model (\autoref{sec:Tagging-model})}
        \label{fig:roberta}
    \end{subfigure}
    \hfill
    \begin{subfigure}[b]{0.3\textwidth}
         \centering
         \includegraphics[page=1, trim=13.0cm 0.0cm 13.0cm 0.0cm, clip=true, width=1.00\linewidth,]{figures/Model.pdf}
         \caption{MLM model (\autoref{sec:MLM-model})}
         \label{fig:mask}
    \end{subfigure}
    \hfill
    \begin{subfigure}[b]{0.3\textwidth}
         \centering
         \includegraphics[page=1, trim=2.0cm 0.0cm 24.0cm 0.0cm, clip=true, width=1.00\linewidth,]{figures/Model.pdf}
         \caption{TDA models (\autoref{sec:TDA-models})}
         \label{fig:topology}
    \end{subfigure}
    \caption{
        Our three main architectures for dialogue term detection.
        Their main distinction is the type of features expected as input. 
        Blue denotes trainable model components. 
        For illustration purposes, here $n=6$.
        \label{fig:three graphs}
    }
\end{figure*}

\section{Background on TDA}
\label{sec:TDA_background}

\emph{Topological data analysis} (TDA) is an emerging toolkit of mathematical methods for analysing the `shape' of data.
In our case, we study point clouds resulting from word vector embeddings, but these general methods apply equally well to spaces of sensor data, images, or audio. 
\emph{Topology} measures important features of a geometric space which are invariant under certain structure preserving transformations such as scaling, rotation, stretching and bending.
\emph{Homology} quantifies the presence or absence of $d$-dimensional \emph{holes} in a geometric space:
In dimension $d = 0$ the homology group $H_{0}$ computes the connected components of a space, while in dimension $d = 1$ the group $H_{1}$ describes the non-fillable closed loops in the space.

Consider a discrete point cloud $P \subset \mathbb{R}^{M}$ equipped with a distance such as the Euclidean metric or the cosine distance. 
To apply topological tools to \(P\), we need to turn \(P\) into a geometric space.
One such `geometrization' is the \emph{Vietoris-Rips complex} $\VR_{\varepsilon}$, which produces, for each non-negative filtration parameter \(\varepsilon\), a \emph{simplicial complex}, a certain higher-dimensional generalization of a graph.
To construct \(\VR_{\varepsilon}\), we consider a collection of higher-dimensional balls of radius \(\varepsilon\) centred at the data points. 
As \(\varepsilon\) increases, the balls grow and merge as in \autoref{fig:vietoris_rips_complex}.
Their overlaps determine the vertices, edges, triangles and higher-dimensional pieces of the complex $\VR_{\varepsilon}$.

The motivation for varying $\varepsilon$ is to measure the `scale' or `resolution' of different topological features.
The filtration parameters $\varepsilon$ at which different $k$-dimensional holes appear and disappear in $\VR_{\varepsilon}$ are summarized in a multiset of points in the plane, visually represented as a \emph{persistence diagram} as in \autoref{fig:south_persistence_diagram}.
Each dot in the diagram corresponds to a feature.
Its horizontal coordinate is the birth time, its vertical coordinate the death time of the feature.
The farther a dot is away from the diagonal, the longer the corresponding feature persists across the range of the parameter $\varepsilon$, and thus the more likely it is to reflect a large-scale topological property of the point cloud $P$.
For an overview of persistent homology from a computational perspective, see~\citet{edelsbrunnerharer2010computationaltopology}.

\section{Dialogue Term Detection}
\label{sec:tagging} 

\subsection{Term Tagging}
\label{sec:Tagging-model}




Our ultimate goal is to extract terms describing domains, slots and values from raw dialogues. In order to achieve this, we adopt the BIO-tagging mechanism presented by~\citet{qiu-structure-2022}. 
In the seed corpus, the spans where concepts occur are tagged with labels `B' (beginning of concept), `I' (inside of concept) and `O' (outside of concept), without distinguishing between different concepts. 
The baseline model is trained on RoBERTa~\cite{Liu2019RoBERTaAR} embeddings as features, and shows modest generalization capabilities when tested in leave-one-out domain experiments.

We investigate two fundamentally different feature sets to increase the generalization capability of models fine-tuned for BIO-tagging.
For each feature, we use a specific input projection and train a transformer followed by a token-level classification head. 
This architecture is illustrated in \autoref{fig:three graphs}.
As the models extract different terms depending on the feature type they are trained on, we use the union of the predictions of all the TDA models, respectively, of all the models, to obtain the final set of terms.
One may also build a combined model using all features as joint input, however due to the nature of the training this would maximize accuracy and not recall.

\subsection{MLM Model}
\label{sec:MLM-model}


The first feature set we consider stems from context-level information captured by large pretrained masked language models (MLM) like BERT~\cite{Devlin2019BERTPO} and RoBERTa. 
Our hypothesis is that, based on how confident an MLM is in predicting a certain word, we can infer the meaningfulness of said word.
We introduce the \emph{masked language modelling score} (MLM score)
$s(w_i) = 1 - p_{\textrm{MLM}}(w_i \mid [w_1, \dots, w_{i-1}, \MASK, w_{i+1}, \dots, w_n])$ 
as the probability that the word $w_i$ is not predicted by the MLM for the $\MASK$ token on position $i$, based on the context 
$[w_1, \dots, w_{i-1}, \MASK, w_{i+1}, \dots, w_n]$.
Thus, meaningful words should have a high MLM score, as illustrated in \autoref{sec:appendix:maskscore}.
The total MLM score $s_{w_i}$ of a word is the average of all scores of all appearances of the word in the data-set.

\subsection{TDA Models}
\label{sec:TDA-models}

Topological features allow us to address the following problem observed in transfer learning:
Tagging models trained directly on the word embedding vectors derived from one dialogue data-set do not generalize well to the embeddings of a different data-set.
We propose the investigation of \emph{topological features} of neighbourhoods of word vectors.
Such topological features capture geometric properties that are invariant under distance-preserving transformations of the data points, and are more generalizable and stable under perturbation than the word-vectors and language model features themselves.
Our hypothesis is that these features reflect properties of words that are data-set independent.

The simplest topological feature we examine is a \emph{codensity vector} that measures the data density in neighbourhoods of various sizes of a given word vector.  A second, far more sophisticated feature that we utilize is \emph{persistence}.
As explained in \autoref{sec:TDA_background}, persistence detects geometric features of a data-set at different scales. 
While degree zero persistence is closely related to density measures, higher degree persistence captures more refined information. 
Finally, we also investigate the \emph{Wasserstein norm} as a two-dimensional summary of persistence.
We now describe these topological descriptors with more mathematical rigour.

\begin{figure}
    \centering
    \includegraphics[width=\linewidth]{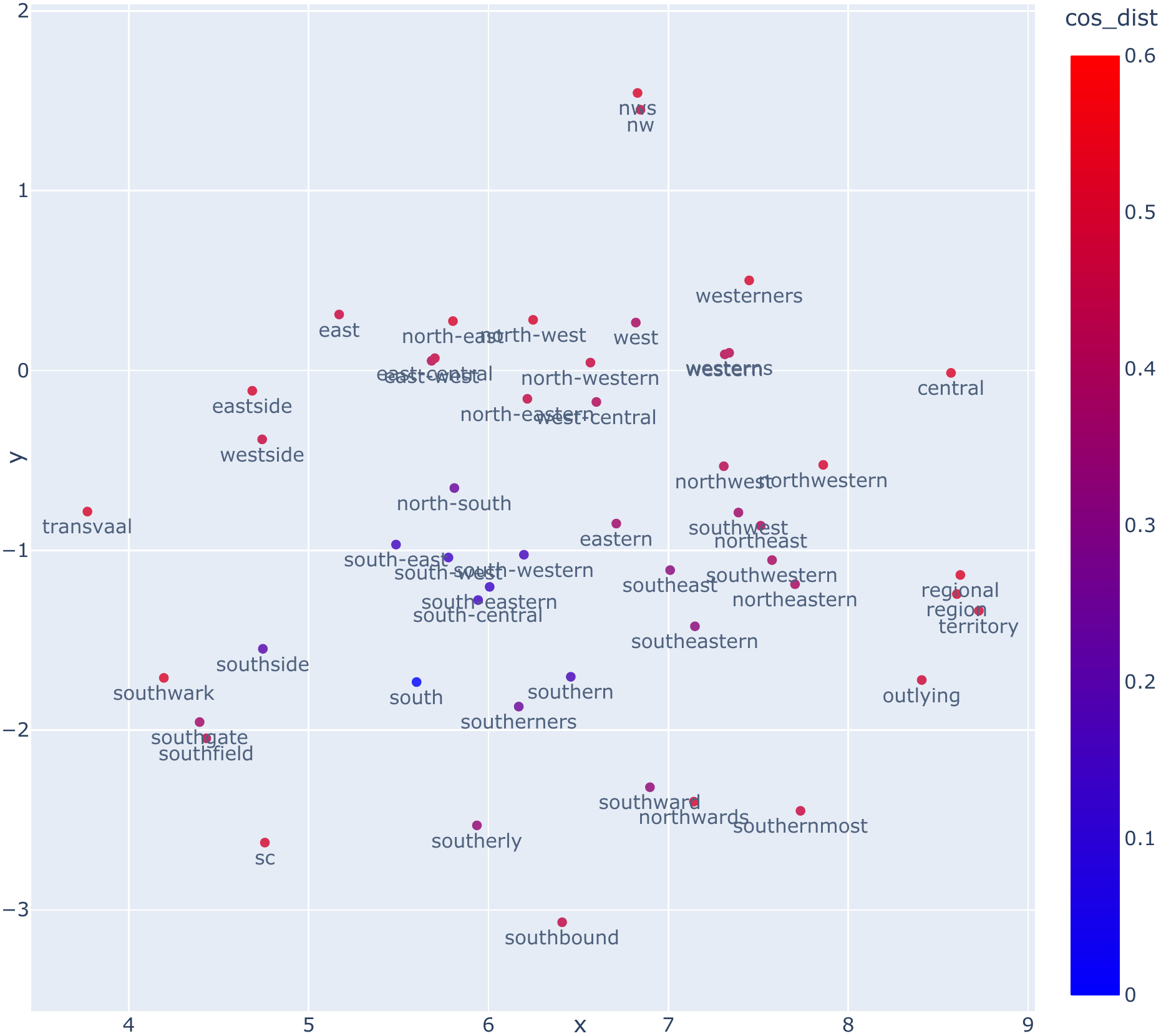}
    \caption{
        2-dim.\ t-SNE projection 
        of the neighbourhood \(\mathcal{N}_{50}(w)\) of \(w =\text{`south'}\).  Colours indicate cosine distance from $w$ in the original $384$-dim.\ embedding space.
        \label{fig:south_tsne_neighborhood}
    }
\end{figure}

\paragraph{Word embedding neighbourhoods}

Our topological tagging models use descriptors derived from neighbourhoods of words in the embedding space.
The neighbourhoods are defined relative to a point cloud $X \subset \mathbb{R}^{M}$ constructed from the word vectors of an ambient vocabulary embedding.
For a given centre $w \in \mathbb{R}^{M}$, let $\mathcal{N}_{n}(w) \subset X \cup \{ w \}$ denote the subset of the $n$ nearest neighbours of $w$ with respect to cosine distance (including \(w\) itself).
In our experiments, we employ neighbourhoods of size $n=50$. See \autoref{fig:south_tsne_neighborhood} or  \autoref{sec:appendix_neighbourhoods} for examples.

The ambient point cloud $X$ in the embedding space needs to be independent of the specific dialogue vocabulary, so that the resulting persistence features of the neighbourhoods remain comparable.
Our vocabulary consists of the 50,000 most common words in the English language, extracted from~\citet{grave2018learning}. 
The embeddings for the point cloud $X$ are created from the SentenceTransformers~\cite{reimers-2019-sentence-bert} \texttt{paraphrase-MiniLM-L6-v2} model.
These dense embeddings of dimension $M = 384$ can be meaningfully compared with cosine similarity.
Note that even though we are using a contextualized model for creating the embeddings, we obtain a `static' point cloud $X$ containing the 50,000 vocabulary vectors.
For building the neighbourhood $\mathcal{N}_{n}(w)$ of a word $w$ not contained in the ambient vocabulary $X$, we first produce $w$'s SentenceTransformers embedding.

\paragraph{Codensity}

The \emph{$k$-codensity} in a point $w$ of a point cloud $P \subset \mathbb{R}^{M}$ is defined as the distance from $w$ to the $k$th nearest point in $P$.
Thus, points with many neighbours at a close distance have a small codensity, which corresponds to a large density of the point cloud around the point $w$.
We construct a 6-dimensional vector $c$ containing the $k$-codensity of $\mathcal{N}_{50}(w)$ at $w$ for $k \in \{ 1, 2, 5, 10, 20, 40 \}$, with the intention of quantifying the neighbourhood density at various scales.

\begin{figure}
    \centering
    \includegraphics[width=0.85\linewidth]{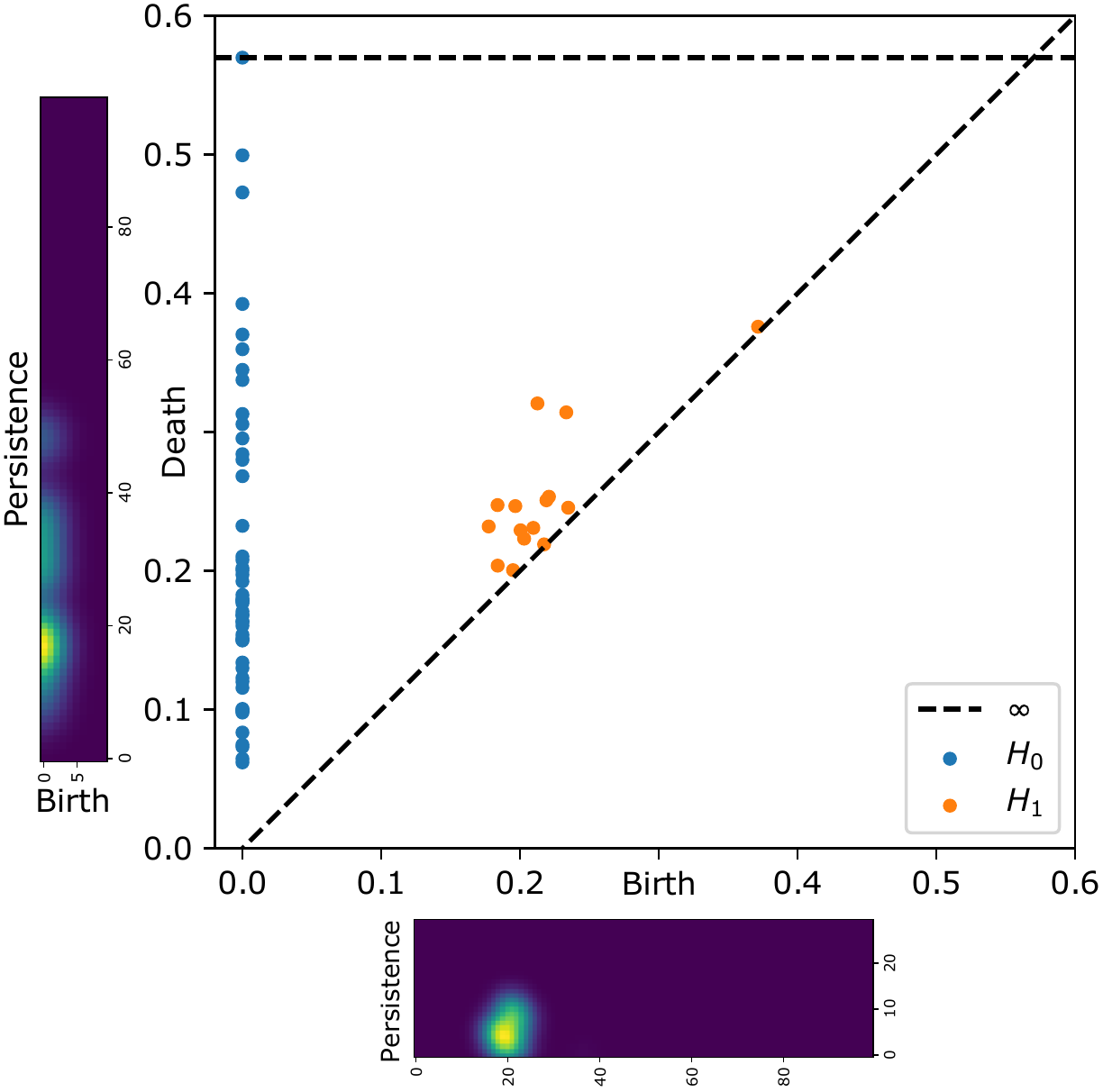}
    \caption{
        Persistence diagram of 
        $\mathcal{N}_{50}(w = \text{`south'})$ for \(H_0\) (blue dots) and \(H_1\) (orange dots) and corresponding persistence images (left: $H_{0}$, bottom: $H_{1}$).
    }
    \label{fig:south_persistence_diagram}
\end{figure}

\paragraph{Persistence}

We produce the persistence diagram (PD) of the sub-point-cloud $\mathcal{N}_{n=50}(w) \subset \mathbb{R}^{384}$ with filtration parameter in the range $[0, 1]$ using cosine distances.
Practically, we apply Ripser~\cite{Bauer2021Ripser} and its Python interface~\cite{ctralie2018ripser} for computations of $H_{0}$ and $H_{1}$ with $\mathbb{F}_{2}$-coefficients.
We restrict to 0- and 1-dimensional homology to keep the computational costs reasonable.
The resulting persistence diagram is a multiset of points in the unit square $[0,1]^{2}$, as in \autoref{fig:south_persistence_diagram}.

Before we can pass the persistence diagrams into the tagging model, we have to apply a \emph{vectorization} step, i.e., map the persistence diagrams into a space which is suitable for training machine learning classifiers.
For this we use \emph{persistence images}~\cite{adams2017persistenceimages}, a short overview of the construction and our choice of parameters is given in \autoref{appendix:persistence_diagram_vectorization}.
\autoref{fig:south_persistence_diagram} contains an example of the persistence images for the `south' neighbourhood.

\paragraph{Wasserstein norm}

The \emph{Wasserstein distance} is a commonly applied measure of similarity of persistence diagrams~\cite{cohensteineredelsbrunner2010Lpstablepersistence}.
In our case, it is a rough numerical estimate of the similarity of the shapes of neighbourhoods.
The \emph{Wasserstein norm} $\norm{D}$ is the Wasserstein distance from $D$ to the empty diagram.
For constructing the input features of the Wasserstein models, we compute the order-$1$ Wasserstein distances with Euclidean ground metric using the GUDHI library~\cite{gudhi:urm} separately for the $H_{0}$ and $H_{1}$ persistence diagrams, leading to a 2-dimensional Wasserstein input vector $\omega$.

\subsection{Training \& Inference}

The MLM score model (\autoref{fig:mask}) and the TDA models (\autoref{fig:topology}) use the following input projections of the respective input features:
The $100$-dimensional $H_{0}$ persistence image vector and $30 \times 100$-dimensional $H_{1}$ persistence image are passed into the model independently and concatenated after downscaling $H_{1}$ to dimension $396$ via a convolutional layer with kernel size $35 \times 25$.
Then they are input to a transformer with hidden dimension $h=496$ and $8$ attention heads.
The transformer output is the input for a token-level classification head after passing through a dropout layer.
\newline
The $6$-dimensional codensity vector $c$, the $2$-dimensional Wasserstein norm vector $\omega$ and the single-dimensional MLM score $s$ are all upscaled to hidden dimension $h = 128$ via a 2-layer fully connected neural network to expand the representation space, before being put into three separate transformers with hidden dimension $h = 128$ and $16$ attention heads.
The transformer sequence output passes through a dropout layer into the token-level classification head.
The token-level classification head consists of a dropout layer, a feed-forward layer with hidden dimension $h$, another dropout, $\tanh$ for activation and an output projection to dimension $3$ corresponding to the three possible BIO tags. 
The classification head is based on the implementation in the HuggingFace library~\cite{huggingfaceref}, where the dropout rate for all layers is $0.1$.

We utilize RoBERTa encoders in two of our models (see \autoref{fig:three graphs}), once to obtain MLM scores with fixed parameters, and once to obtain contextual semantic embeddings after fine-tuning on the BIO-tagging task.
We train each model on MultiWOZ with cross-entropy loss and a learning rate of $4\mathrm{e}{-5}$ using the AdamW optimizer~\cite{loshchilov2018decoupled}, warm-up for $10\%$ of total training steps and linear decay afterwards. 
We train for 15 epochs, with training stopping early if the loss on the validation set stays within a range of $\delta=0.005$ and batch size 128 on one NVIDIA Tesla T4 GPU. 
For the much smaller training data in the leave-one-out experiments, the batch size is decreased to 32.

\section{Experiments}
\label{sec:experiments}

We conducted experiments to answer the following questions: 
(1) Is it possible to train a model on the seed data-set that achieves a high recall rate on the unseen ontology? 
(2) Which of the proposed features is most valuable for that purpose? 
(3) What kind of concepts is the model able to find?

Note that we are mainly focusing on recall as evaluation measure, while retaining the F1-score of the baseline model.
Improvements in precision can be achieved with further post-processing, such as clustering~\cite{qiu-structure-2022,yu-unsupervised-schema-2022}. 

\subsection{Data-sets}

We use two well-established data-sets for modelling task-oriented dialogues. 
MultiWOZ~\cite{budzianowski2018multiwoz,eric2019multiwoz} is a corpus of human-to-human dialogues that were collected in a Wizard-of-Oz fashion.
Each conversation has one or more goals that revolve around seeking information about or booking tourism-related entities.
The data-set consists of over 10,000 dialogues covering 6 domains. 
There are 30 unique domain-slot pairs that take approximately 4,500 unique values.
Value occurrences are annotated with span labels. 
MultiWOZ is the seed set for training all of our term extraction models.

The Schema-Guided Dialogue (SGD) data-set~\cite{rastogi2020towards} is considerably larger than MultiWOZ, with dialogues spanning across 20 domains that represent a wide variety of services. 
The number of unique values is almost four times larger than in MultiWOZ. 
This means that any model trained on the significantly more narrow MultiWOZ seed data would need to be able to generalize extremely well to achieve reasonable term extraction performance on SGD.
Therefore, SGD is an ideal data-set for our zero-shot experiments.

\begin{figure}[t]
    \centering
	\includegraphics[page=4, trim=6.35cm 6.1cm 6.3cm 6.2cm, clip=true, width=1.00\linewidth,]{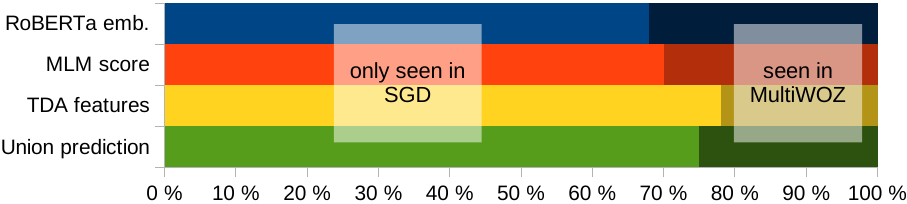}
	\caption{
	    Percentage of extracted terms which were already seen during training or are only seen on SGD during test time.
	    \label{fig:seen_on_woz}
	}
\end{figure}

\subsection{Set-up}

In order to investigate the models' ability to extract terms in an unseen domain, we design two experiments. 
First, we conduct a leave-one-out domain experiment on MultiWOZ, similar to the approach taken by~\citet{qiu-structure-2022}, with two important differences. 
We focus mainly on recall as the adequate evaluation measure for term extraction, and we do not allow partial matches of the tagged term.
When designing the matching function, we were guided by the tolerance threshold of a picklist-based dialogue state tracker.
For example, the term extractor is allowed to match `an expensive' with the golden term `expensive', as having a non-content word in the term would make no difference to the tracker. 
However, matching `Pizza Hut' with the golden term `Pizza Hut Cherry Hinton' is considered a false positive, as `Pizza Hut' would not be precise enough for the tracker to distinguish entities.
Note that such matches were considered by~\citet{qiu-structure-2022} as true positives, so our matching function is stricter.  
For both training and testing we limit ourselves to user utterances, as the system utterances may contain API calls, which is already structured data.

For the second experiment, we train our models on the training portion of the MultiWOZ data-set and test it on the SGD data-set.
We then examine the overlap in true positives between models using different features. We also analyse the models' abilities to extract terms referring to different domains and slots, highlighting easy and difficult terms.


\subsection{Results}
\label{sec:results}

\paragraph{Leave-one-out domain}
\begin{table}[t]
    \setlength{\tabcolsep}{3pt}
	\centering
	\small
	\begin{tabular}{@{}llccccc@{}} 
		\toprule
		Approach & Measure & Taxi & Rest. & Hotel & Attr. & Train \\
		\midrule
		\multirow{2}{*}{RoBERTa} & F1 & 0.87 & 0.81 & 0.68 & 0.91 & 0.84 \\ 
		\multirow{2}{*}{embeddings} & Recall & 0.87 & 0.89 & 0.95 & 0.94 & 0.92 \\
		& Precision & 0.87 & 0.76 & 0.53 & 0.89 & 0.77 \\ 
		\midrule
		\multirow{2}{*}{MLM} & F1 & 0.44 & 0.47 & 0.32 & 0.42 & 0.57 \\
		\multirow{2}{*}{score} & Recall & 0.43 & 0.48 & 0.69 & 0.53 & 0.72 \\ 
		& Precision & 0.46 & 0.46 & 0.21 & 0.35 & 0.47 \\
		\midrule
		Persistence & F1 & 0.72 & 0.61 & 0.41 & 0.63 & 0.65 \\ 
		image & Recall & 0.79 & 0.69 & 0.87 & 0.65 & 0.92 \\ 
		vectors & Precision & 0.67 & 0.54 & 0.27 & 0.61 & 0.50 \\ 
		\midrule
		& F1 & 0.57 & 0.46 & 0.38 & 0.51 & 0.62 \\ 
		Codensity & Recall & 0.51 & 0.48 & 0.64 & 0.59 & 0.76 \\ 
		& Precision & 0.64 & 0.44 & 0.27 & 0.45 & 0.52 \\ 
		\midrule
	    \multirow{2}{*}{Wasserstein} & F1 & 0.57 & 0.50 & 0.45 & 0.46 & 0.48 \\ 
		\multirow{2}{*}{norm} & Recall & 0.58 & 0.53 & 0.46 & 0.51 & 0.69 \\ 
		& Precision & 0.57 & 0.47 & 0.45 & 0.43 & 0.37 \\ 
		\midrule
		
		\multirow{2}{*}{TDA} & F1 & 0.65 & 0.53 & 0.33 & 0.52 & 0.47 \\
		\multirow{2}{*}{features} & Recall & 0.84 & 0.81 & 0.89 & 0.84 & 0.94 \\ 
		& Precision & 0.53 & 0.39 & 0.20 & 0.37 & 0.31 \\ 
		\midrule
		\multirow{2}{*}{Union} & F1 & 0.65 & 0.53 & 0.26 & 0.49 & 0.44 \\
		\multirow{2}{*}{prediction} & Recall & \textbf{0.95} & \textbf{0.92} & \textbf{0.97} & \textbf{0.98} & \textbf{0.98} \\
		& Precision & 0.50 & 0.37 & 0.15 & 0.33 & 0.28 \\
		\bottomrule
	\end{tabular} 
	\caption{Leave-one-out results on MultiWOZ.}
	\label{table:leaveoutdom}
\end{table}

We remove one of the five MultiWOZ domains in training and only test on it, so the model has not seen any dialogues in the left-out domain. 
We only utilize single domain dialogues in the training and test set.
Results in \autoref{table:leaveoutdom} show that the recall increases for each unseen domain experiment when adding the predictions by the models trained on persistence and language modelling features to form the union prediction.

\paragraph{Unseen ontology}
\begin{table}[t]
    \centering
    \small
	\begin{tabular}{@{}lcccc|c@{}}
		\toprule
		Approach & F1 $\uparrow$ & Rec. $\uparrow$ & Prec. $\uparrow$ & L2 $\downarrow$ & Tags \\ 
	    \midrule
		RoBERTa emb. & 0.45 & 0.35 & 0.63 & 0.29 & 2757 \\
		MLM score & 0.34 & 0.34 & 0.35 & 0.35 & 4933 \\
		PI vectors & 0.47 & 0.46 & 0.48 & \textbf{0.20} & 4775\\ 
		Codensity & 0.37 & 0.34 & 0.42 & 0.52 & 4054\\ 
		Wasserst. n. & 0.42 & 0.40 & 0.44 & 0.62 & 4536\\
		TDA features & 0.48 & 0.63 & 0.39 & - & 8189 \\
		Union pred. & \textbf{0.48} & \textbf{0.74} & 0.36 & - & 10398 \\
		\bottomrule
	\end{tabular} 
	\caption{
	    Dialogue term extraction results on SGD with models trained on MultiWOZ together with the total number of tagged terms per model. There are 5008 target terms in SGD. 
	    L2-norm is used as uncertainty measure for the single models.
		\label{table:sgd_results}
	}
\end{table}

The results in \autoref{table:sgd_results} show that adding the predictions of the new feature models improve both recall and F1-score significantly for term extraction on the unseen SGD ontology compared to the language model only baseline, without the need to fine-tune the embeddings on the token classification task with any SGD data. 
In \autoref{fig:seen_on_woz} the percentage of completely new terms found in the predictions of each model is shown.
The TDA feature model predictions contain mostly unseen terms.
Confidence scores would be critical in a subsequent automatic ontology construction.
We compare the L2-norm of the model's predictions to the ground truth label, showing that the model trained on persistence image vectors from MultiWOZ has the highest confidence score on the unseen SGD data.

\begin{table*}[t]
  \centering
  \small
	\begin{tabular}{@{}p{4.0cm}p{2.9cm}p{4.8cm}p{2.9cm}@{}}
		\toprule 
		Seen in MultiWOZ & Only seen in SGD & False negatives & False positives  \\ 
		\midrule
		Lebanese; Hotel Indigo London-Paddington; LAX International Airport; The Queen's Gate Hotel; Hair salon  & Delta Aesthetics; McDonald's; 3455 Homestead Road; receiver; Pescatore
		& Little Hong Kong; Yankees vs. Rangers; Dr. Eugene H. Burton III; 341 7th street; La Quinta Inn by Wyndham Dacramento Downtown
		& Especillay by; Bears vs; Angeles and; Polk Street; theater please; resrevation; neaarby \\ 
		\bottomrule
	\end{tabular} 
	\caption{
	    Example predictions of the Union model on SGD (typos are reproduced as they appear in the data-set).
	    Examples for each of our other models can be found in \autoref{sec:appendix:other_examples}.
	    \label{table:examples_seen_and_negative}
	}
\end{table*}

\begin{table*}[t]
  \centering
  \small
	\begin{tabular}{@{}p{3.0cm}p{9.0cm}@{}}
		\toprule 
		& i ' d like to find a steakhouse that ' s not very costly to eat at . \\
		\midrule
		RoBERTa embeddings
		& \phantom{i ' d like to find a} \mybox[fill=blue!20]{steakhouse} \phantom{that ' s} \mybox[fill=blue!20]{not} \phantom{very costly to eat at .} \\
		MLM score
		& \phantom{i} \mybox[fill=blue!20]{'} \phantom{d like to find a} \mybox[fill=blue!20]{steakhouse} \mybox[fill=blue!20]{that} \phantom{' s not very costly to eat at .} \\
		TDA features
		& \phantom{i ' d like to find a} \mybox[fill=blue!20]{steakhouse} \phantom{that ' s not very} \mybox[fill=blue!20]{costly} \phantom{to eat at .} \\
		\bottomrule
	\end{tabular} 
	\caption{
	    Example of a normalized, tokenized utterance together with terms extracted by the different models.
	    Unconnected boxes indicate separate terms, i.e., here the MLM score model assigned a B tag to 'steakhouse' and a B tag to 'that'.
	    More example utterances can be found in \autoref{sec:appendix:other_examples}.
	    \label{table:example_utterance_and_tags}
	}
\end{table*}

\begin{figure}[t]
    \centering
	\includegraphics[width=0.95\linewidth]{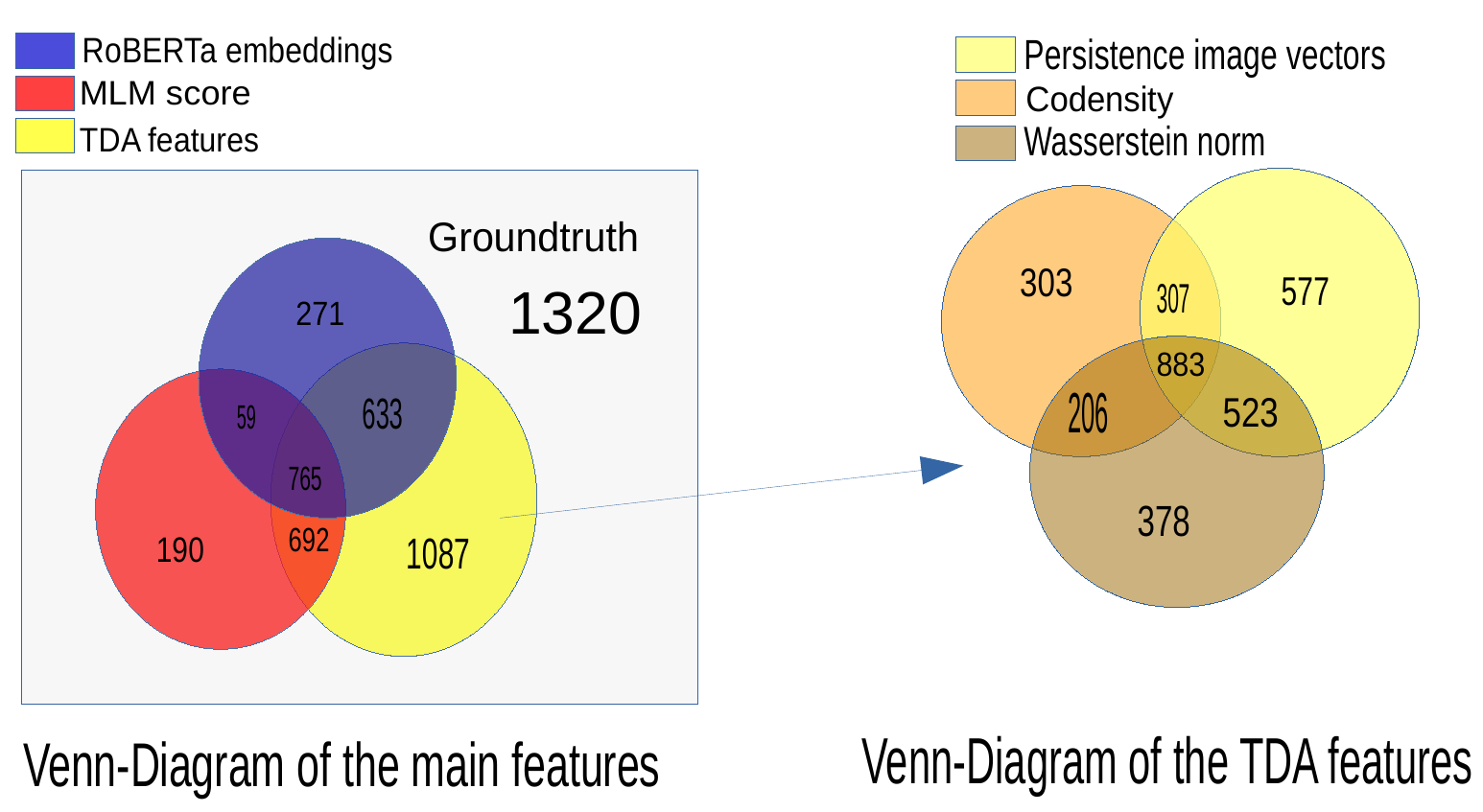}
	\caption{
		Venn-Diagram of SGD terms found in each of the three models using RoBERTa, MLM score, TDA features, as well as analysis of term overlap of the models trained on different TDA features.
		\label{fig:venn_diag_overlap}
	}
\end{figure}

\paragraph{Overlap}

\autoref{fig:venn_diag_overlap} shows that the sets of extracted terms differ significantly by model. 
Therefore, the union of predictions is useful for capturing as many relevant terms as possible. 
The MLM score model already adds more terms to the fine-tuned language model. 
The topological features, however, by far supply the biggest portion of new terms. 
Among the different TDA features, the persistence images yield the largest number of additional terms.

\paragraph{Domain and slot coverage}

\begin{figure}[t]
	\centering
	\includegraphics[page=1, trim=7.0cm 3.5cm 7.5cm 3.6cm, clip=true, width=1.00\linewidth,]{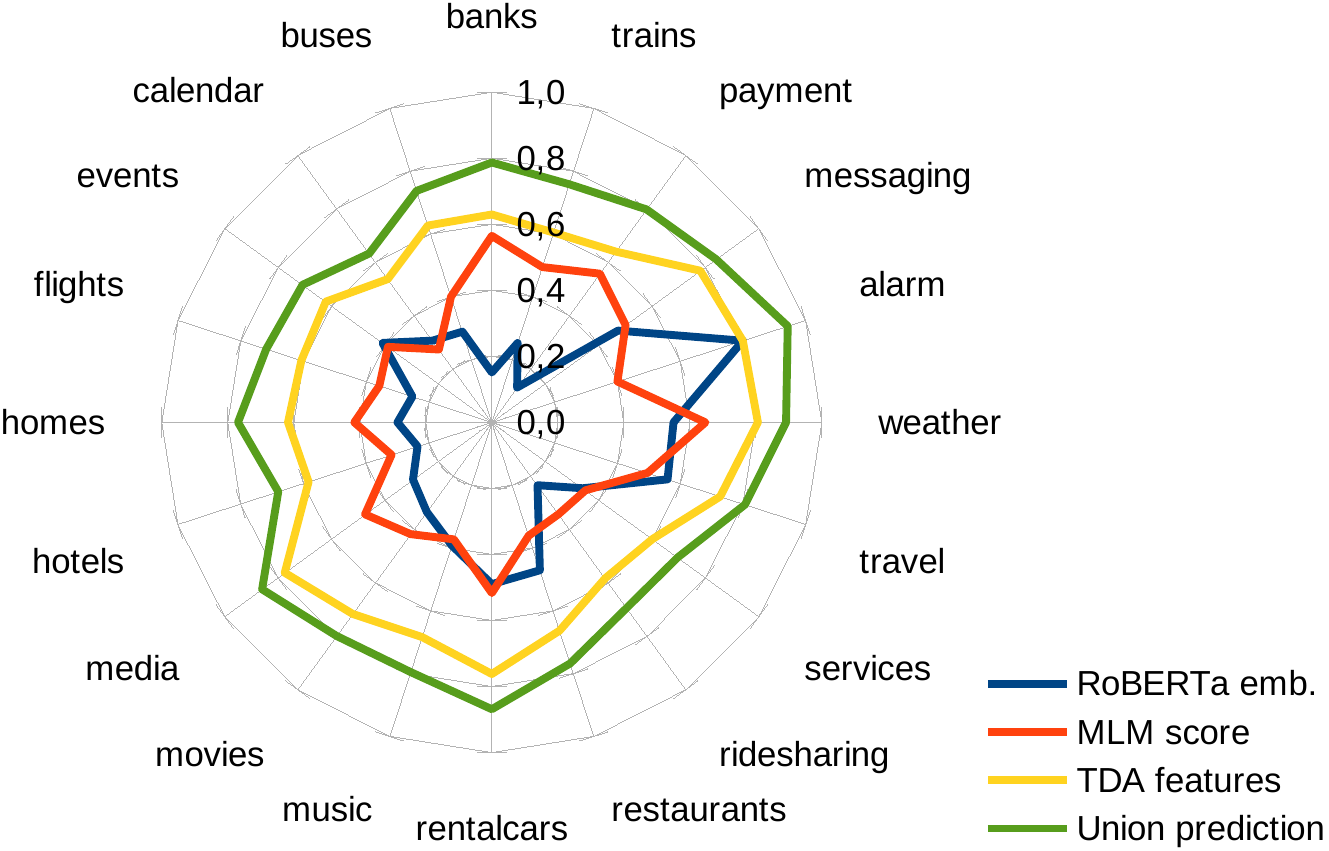}
	\caption{
		Recall per domain on SGD by our models compared with the baseline fine-tuned RoBERTa model.
		\label{fig:per_domain_rec}
	}
\end{figure}

\autoref{fig:per_domain_rec} demonstrates that the different models find various amounts of terms depending on the domain. 
The recall of the TDA models is the highest across all domains, while RoBERTa is only able to outperform the MLM score model in terms of recall in 5 out of 20 domains, e.g., in `music' and `restaurants', which contain many multi-word terms.

\paragraph{Examples}

False negatives tend to be long multi-word terms, as exemplified in \autoref{table:examples_seen_and_negative}.
False positives predominantly include typos and incomplete terms.
Predictions by RoBERTa contain $2.0$ words on average.
In contrast, the MLM score model and TDA feature model term predictions have an average length of $1.6$ and $1.8$ words, respectively.
We give an illustrative instance of terms extracted by the different models from an example utterance in \autoref{table:example_utterance_and_tags}.

\section{Discussion and Future Outlook}
\label{sec:discussion}

Our novel term extraction approach based on topological data analysis and masked language modelling scores significantly outperforms the word-embedding-based baseline on the recall rate both in leave-one-out experiments and when applied to a completely different corpus.
Importantly, our results demonstrate a strong ability of topological data analysis to extract domain independent features that can be used to analyse unseen data-sets.
This finding warrants further investigation.

Our approach still produces a significant number of false positives.
The next step in the ontology construction pipeline, clustering, could be deployed to significantly reduce that number, as has already been demonstrated by~\citet{yu-unsupervised-schema-2022}.
We believe that their approach and our approach could be combined, but that goes beyond the scope of this work. 

However, ultimately, precision is only of secondary importance.
In a typical goal oriented system, we have a dialogue state tracker tracking concepts through conversation. 
Whether or not the tracker is tracking some irrelevant terms does not impact the overall performance of a dialogue system.
All that matters is that the tracker does track every term that actually is a concept.
Of course, the computational complexity of the tracker increases linearly with the number of tracked terms~\cite{heck2020trippy, van-niekerk-2020-setsumbt, lee-2021-t5dst}.
But, as can be seen from \autoref{table:sgd_results}, our method merely doubles the number of terms, so the computational price tag is low.
With this in mind, it is also conceivable that the tracker itself could be utilized to increase the precision. This would be an interesting direction for further research.
%

Some simpler options for improvement are more immediate:
Here, we utilize SentenceTransformers only to provide static embeddings for each word, but of course a similar analysis can be applied to contextualized word embeddings, at the expense of higher computational complexity.
Further, persistence images (\autoref{sec:TDA-models}) could be replaced by features tailored to downstream tasks, such as features obtained from the novel Persformer model~\cite{DBLP:journals/corr/abs-2112-15210_persformer}.

\section{Conclusion}
\label{sec:conclusion}

To the best of our knowledge, we present the first application of topological features in dialogue term extraction. 
Our results show that these features distinguish content from non-content words, in a way that can be generalized from a training domain to unseen domains. 
We believe that these findings are only the tip of the iceberg, and warrant further investigation of topological features in NLP in general. 
In addition, we have shown that masked language modelling scores are useful for term extraction as well. 
In combination, the features we investigate allow us to make a significant step towards automatic ontology construction from raw data.

\section*{Acknowledgements}


RV is supported by funds from the European Research Council (ERC) provided under the Horizon 2020 research and innovation programme (Grant agreement No.\ STG2018 804636) as part of the DYMO project.
CVN and MH are supported by funding provided by the Alexander von Humboldt Foundation in the framework of the Sofja Kovalevskaja Award endowed by the Federal Ministry of Education and Research.
Google Cloud provided computational infrastructure. 
We want to thank the anonymous reviewers whose comments improved the exposition of our paper.

\bibliography{custom}
\bibliographystyle{acl_natbib}

\appendix

\section{Neighbourhoods and Persistence Diagrams}
\label{sec:appendix_neighbourhoods}

\begin{figure*}
    \centering
    \resizebox{.9\linewidth}{!}{
    \begin{tabular}{m{0.65cm} m{8cm} m{8cm} c}
        \toprule
        &
        Neighbourhood &
        Persistence Diagram &
        Wasserst. / Codens. 
        \\ 
        \midrule
        can &
        \includegraphics[width=8cm]{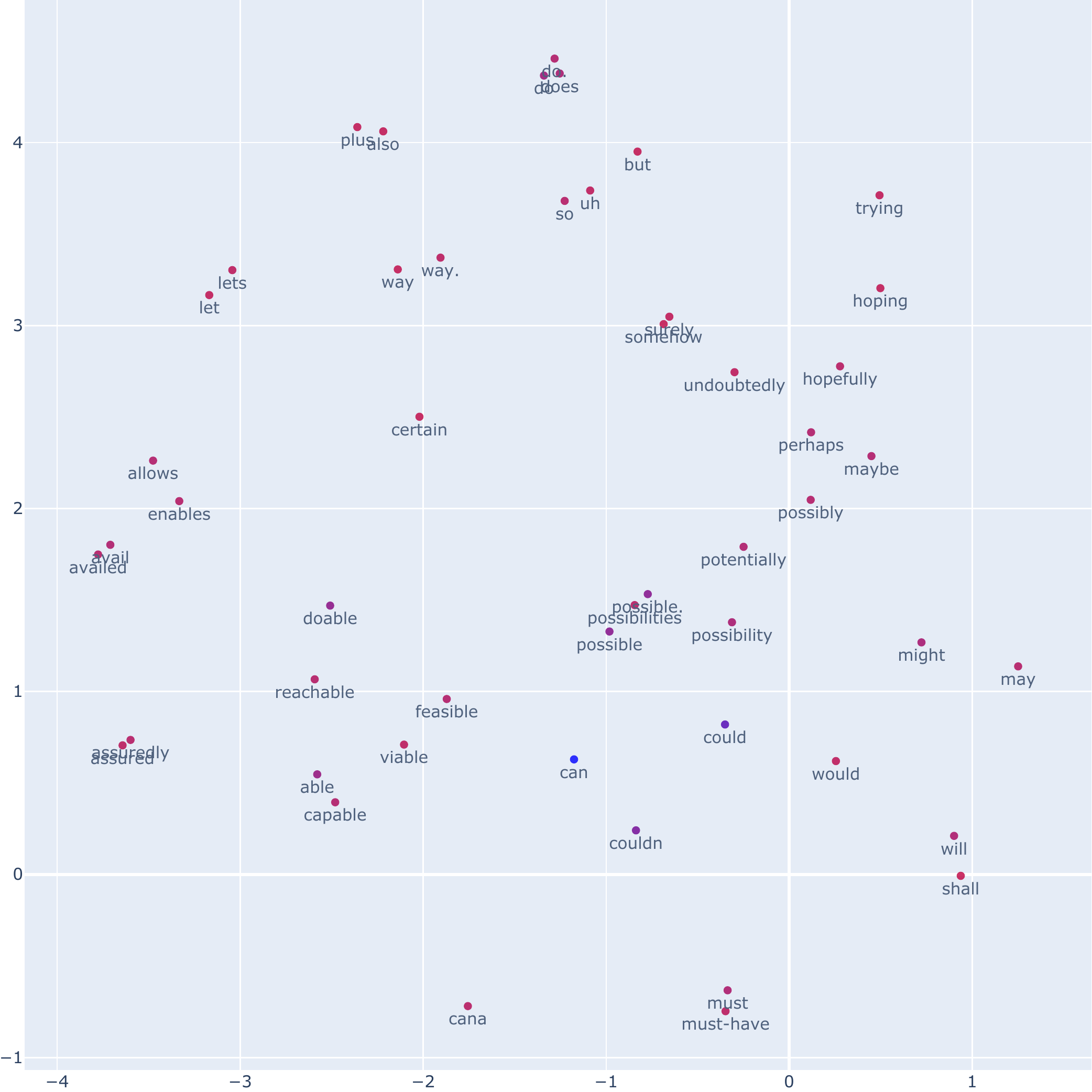}&
        \includegraphics[width=8cm]{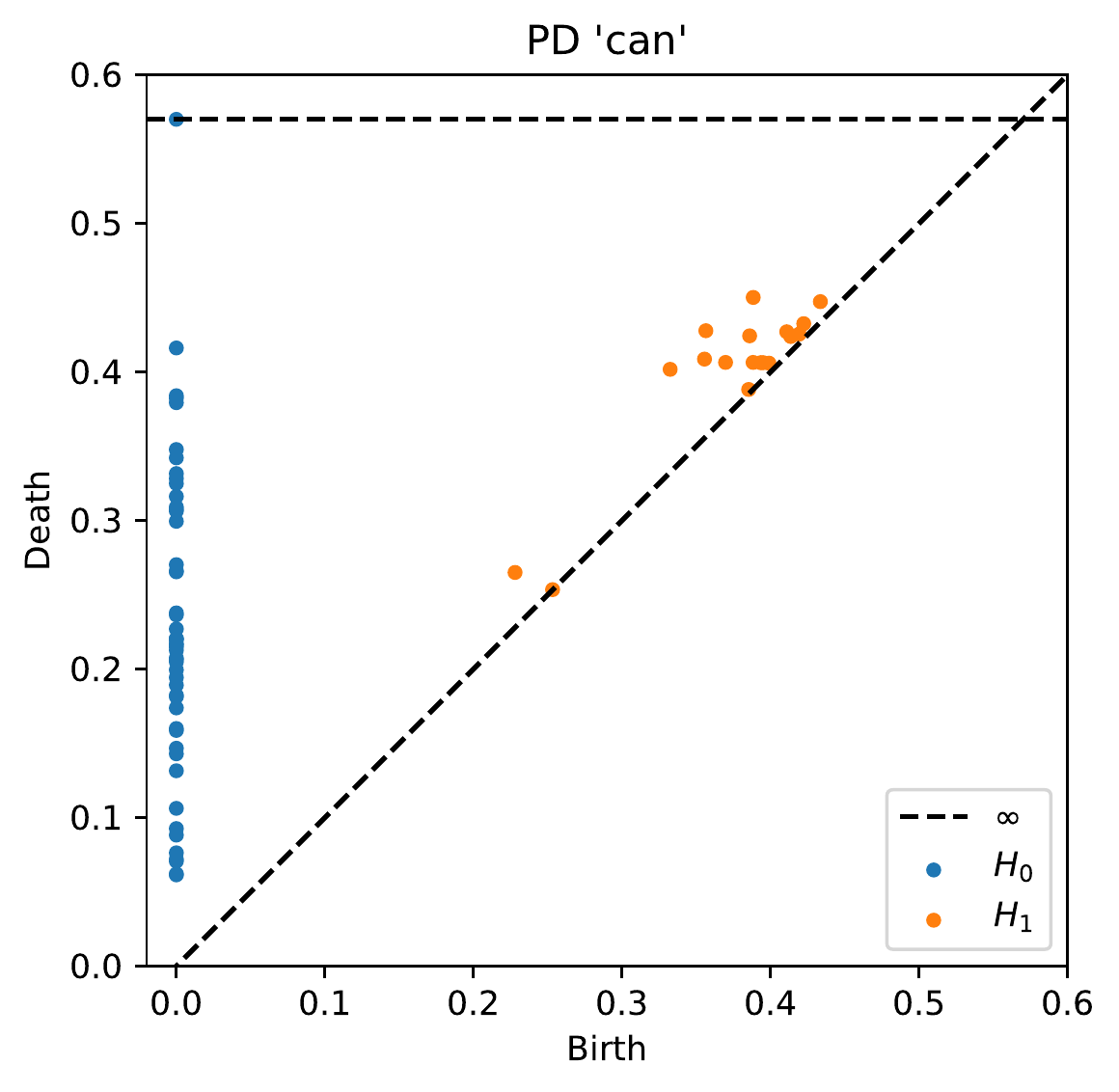}&
        $
        \begin{pmatrix}
            7.712 \\ 0.311
        \end{pmatrix}
        $
        $
        \begin{pmatrix}
            0.181 \\ 0.259 \\ 0.307 \\ 
            0.379 \\ 0.402 \\ 0.439
        \end{pmatrix}
        $
        \\ 
        the &
        \includegraphics[width=8cm]{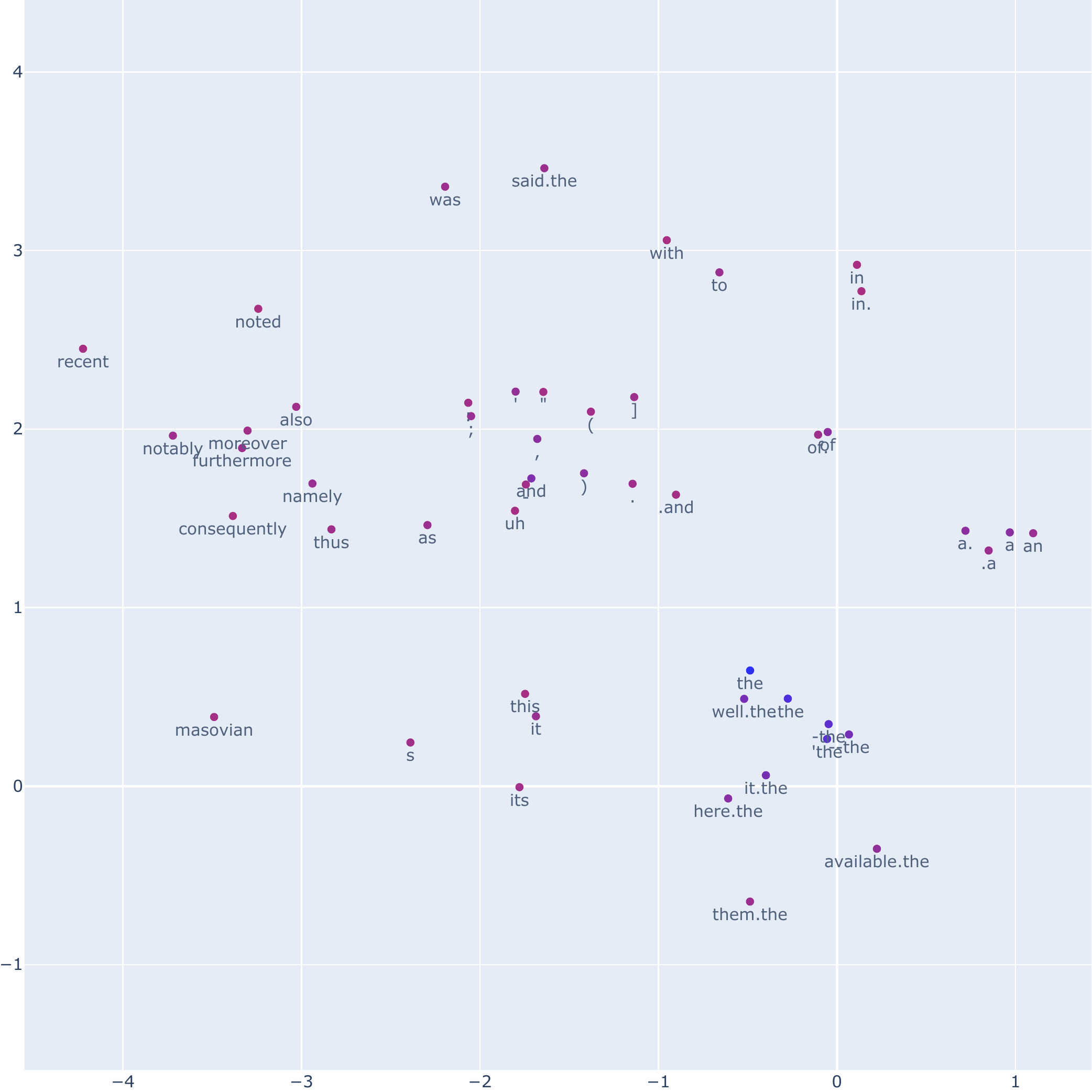}&
        \includegraphics[width=8cm]{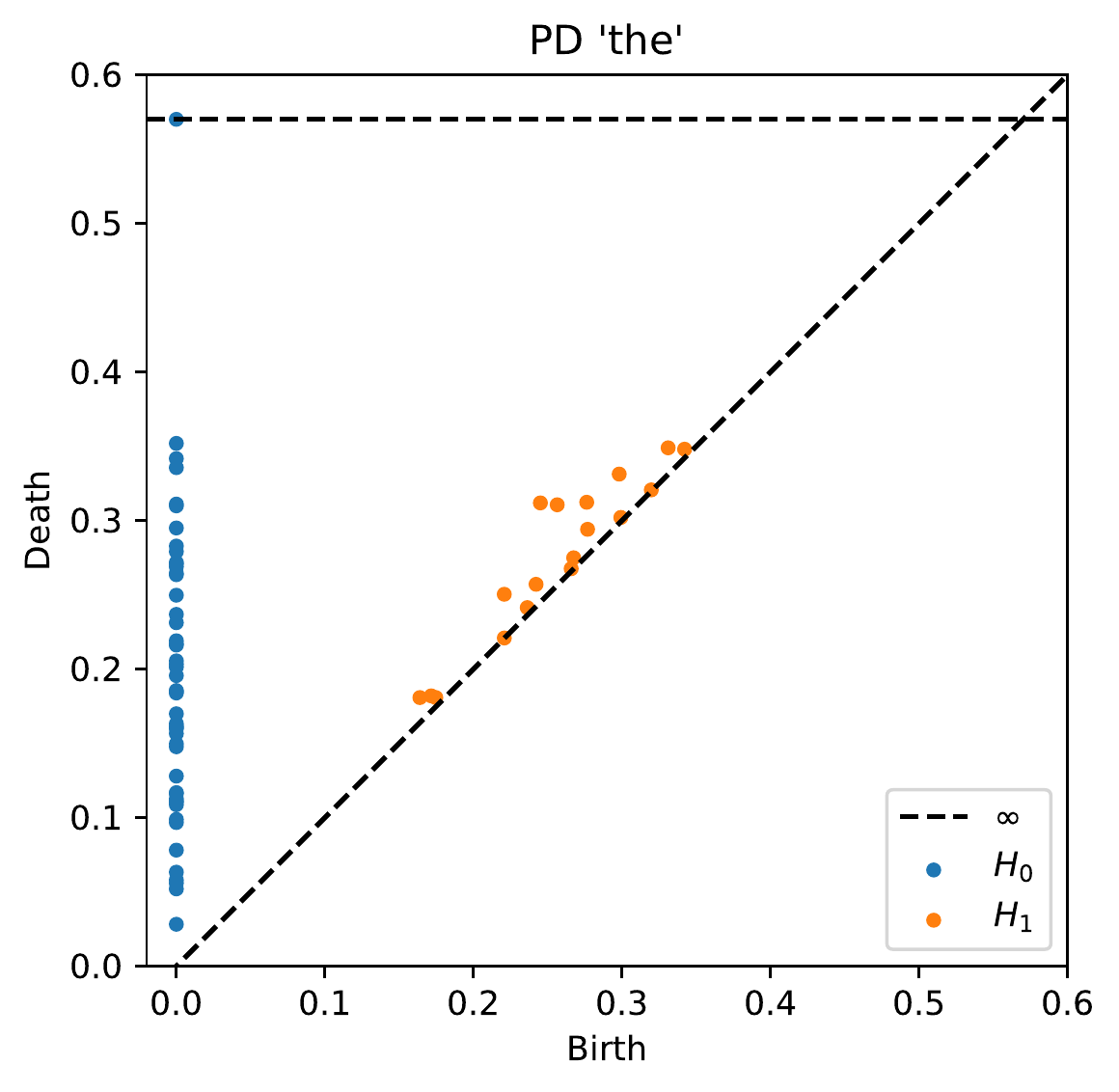}&
        $
        \begin{pmatrix}
            6.490 \\ 0.219
        \end{pmatrix}
        $
        $
        \begin{pmatrix}
            0.099 \\ 0.129 \\ 0.213 \\
            0.275 \\ 0.318 \\ 0.349
        \end{pmatrix}
        $ 
        \\ 
        cheap &
        \includegraphics[width=8cm]{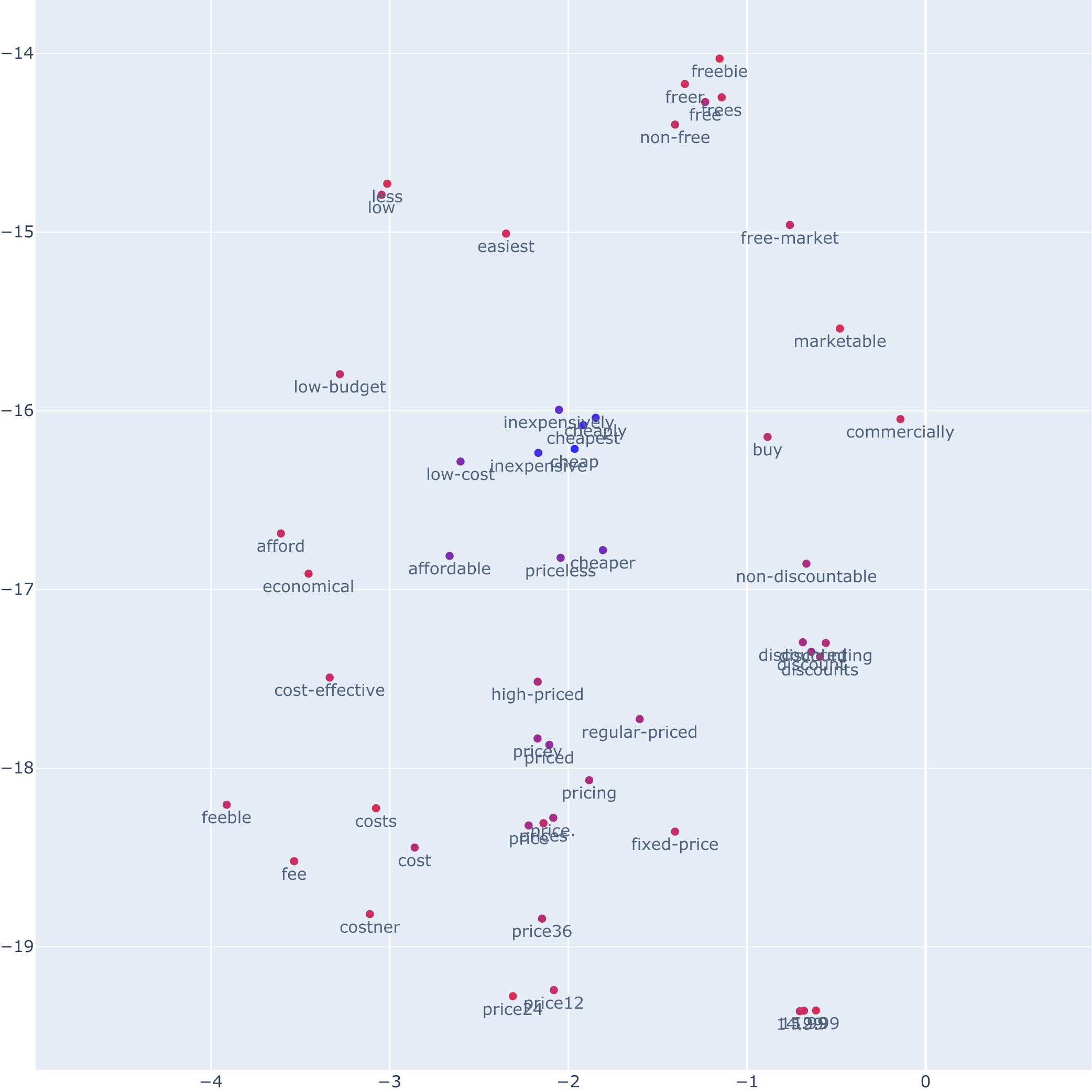}&
        \includegraphics[width=8cm]{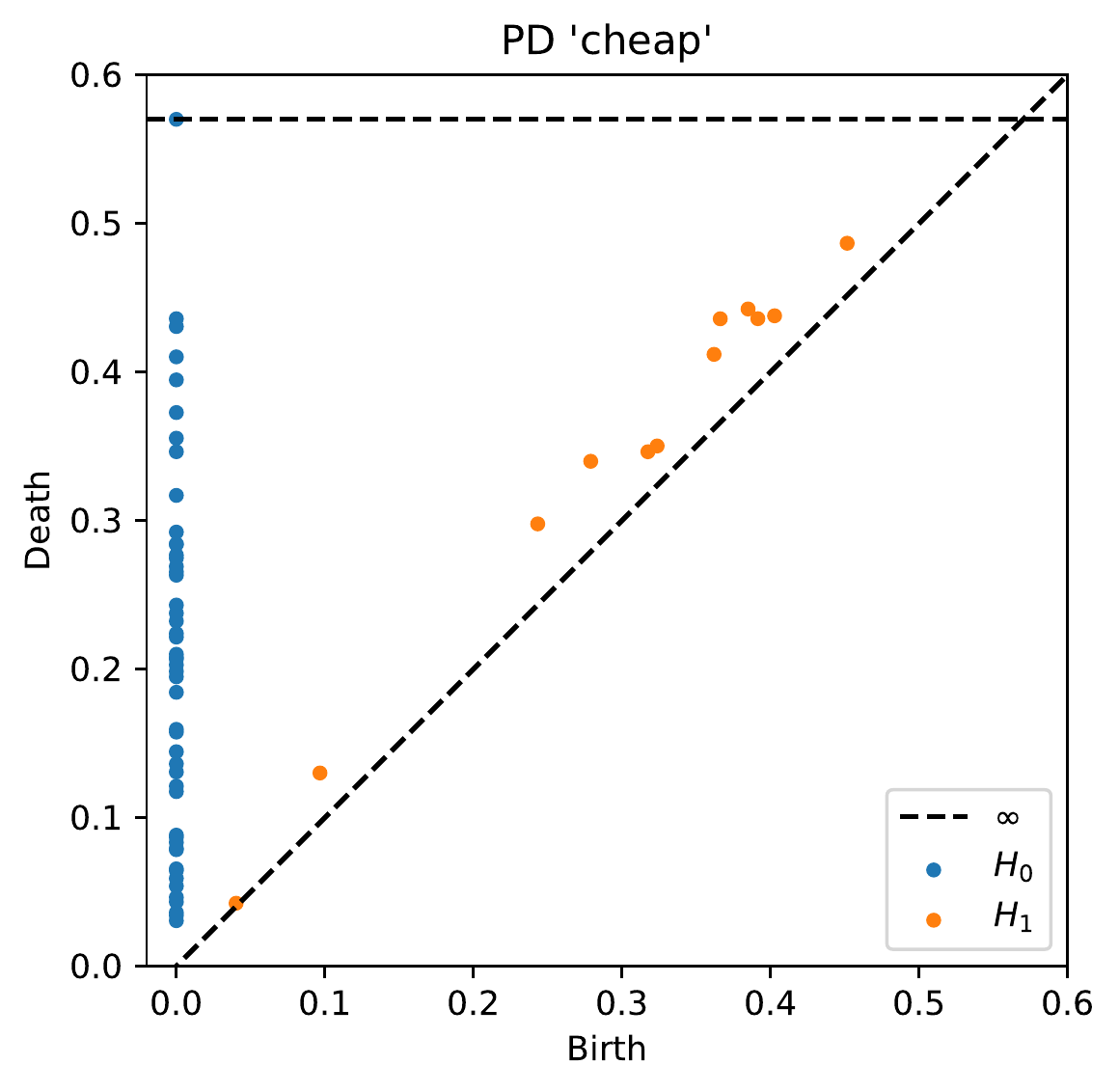}&
        $
        \begin{pmatrix}
            6.823 \\ 0.350
        \end{pmatrix}
        $
        $
        \begin{pmatrix}
            0.059 \\ 0.065 \\ 0.195 \\
            0.322 \\ 0.387 \\ 0.464
        \end{pmatrix}
        $
        \\ 
        hotel &
        \includegraphics[width=8cm]{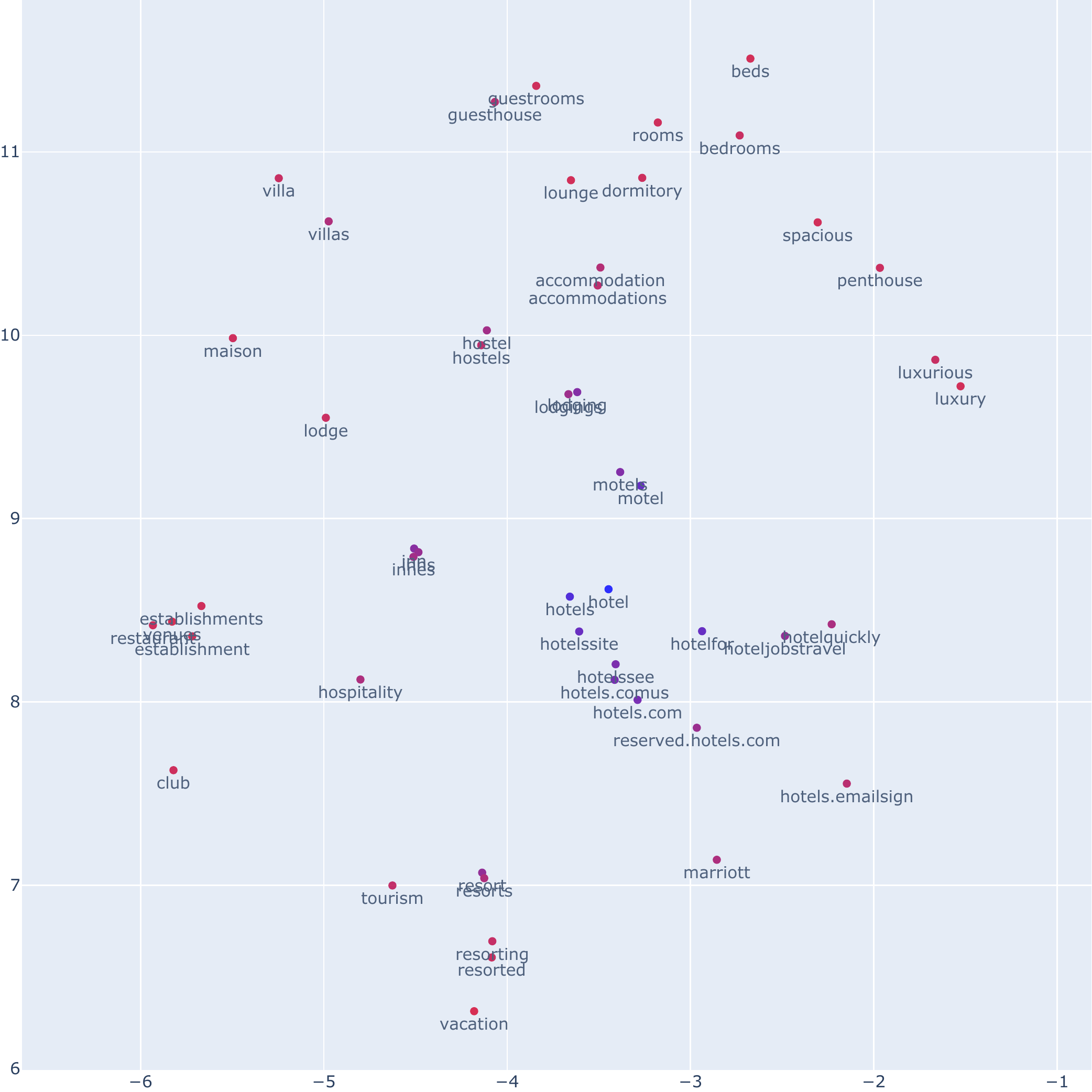}&
        \includegraphics[width=8cm]{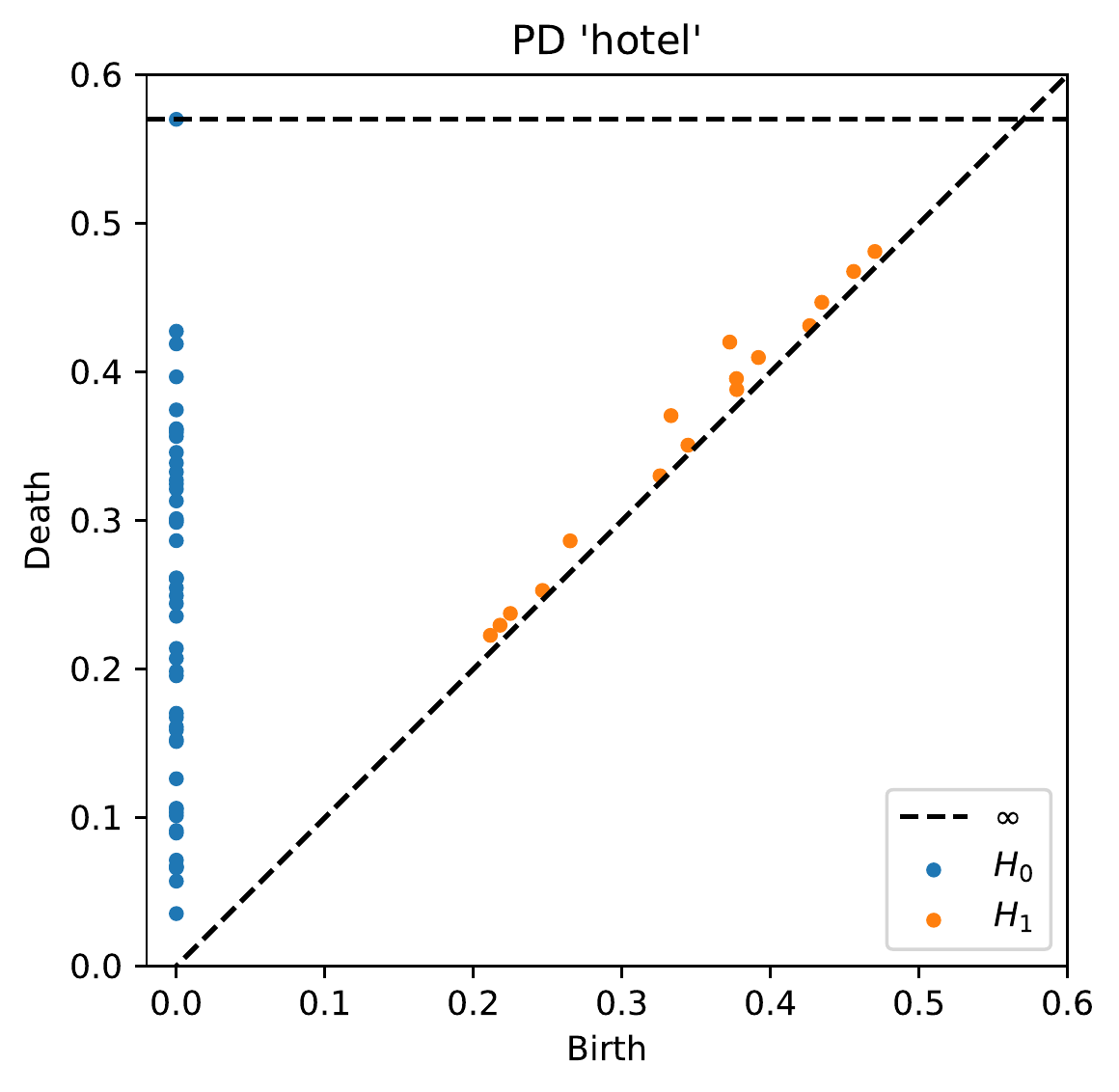}&
        $
        \begin{pmatrix}
            7.902 \\ 0.166
        \end{pmatrix}
        $
        $
        \begin{pmatrix}
            0.106 \\ 0.167 \\ 0.217 \\ 
            0.265 \\ 0.374 \\ 0.473
        \end{pmatrix}
        $
        \\
        \bottomrule
    \end{tabular}
    } 
    \caption{
        2-dimensional t-SNE projection of the neighbourhood $\mathcal{N}_{50}(w)$;
        corresponding Persistence diagram;
        2-dim.\ Wasserstein norm vector (for $H_{0}$ and $H_{1}$);
        6-dim.\ codensity vector (for $k \in \{ 1, 2, 5, 10, 20, 40 \}$).
    }
    \label{fig:table_neighbourhoods_persistence_diagrams}
\end{figure*}

We produce a table with various figures of neighbourhoods, their persistence diagrams, Wasserstein norm vectors and codensity vectors in \autoref{fig:table_neighbourhoods_persistence_diagrams}.

\section{Details about the Persistence Diagram Vectorization Step}
\label{appendix:persistence_diagram_vectorization}

We used the scikit-tda/persim library~\cite{scikittda2019} in the practical implementation of persistence images.

As a first step, the $(\textrm{birth}, \textrm{death})$ coordinates of the dots in the persistence diagram are transformed into $(\textrm{birth}, \textrm{lifetime} = \textrm{death} - \textrm{birth})$ coordinates.
We then place a Gaussian kernel with variance $\sigma = 0.0007$ onto each point in the $(\textrm{birth}, \textrm{lifetime})$ diagram, linearly weighted by the lifetime.
We sum up the various probability distributions and then integrate the resulting function over the patches of a rasterization with a pixel size of $0.1$ of the image plane.
\citet{adams2017persistenceimages} discuss that the performance of the resulting persistence images for downstream tasks is robust in the choices of these parameters.
As usual in the Vietoris-Rips filtration, the birth of all the 0-dimensional homology classes in $H_{0}$ occur for radius $\varepsilon = 0$, and we consider the persistence features in the range $[0.0, 1.0]$.
Thus, we only pass the 0th column of the generated $H_{0}$ persistence image to the model, which is a $100$-dimensional vector.
For the $H_{1}$ persistence image, we take the entire birth range $[0.0, 1.0]$ and persistence range $[0.0, 0.3]$ into account, so that the image has dimensions $100 \times 30$. 

\newpage
\section{Masked Language Modelling Score Examples}
\label{sec:appendix:maskscore}

In \autoref{table:maskscore_examples} the MLM scores on MultiWOZ and SGD of example words show that the score is high for meaningful words across data-sets.


\begin{table}[h!]
	\centering
	\small
	\begin{tabular}{@{}lcc@{}}
		\toprule
		Word & Score on MultiWOZ & Score on SGD \\
		\midrule
		cheap & 0.96 & 0.92 \\
		restaurant & 0.86 & 0.86 \\
		the & 0.59 & 0.63 \\
		how & 0.70 & 0.67 \\
		not & 0.45 & 0.50 \\ 
		\bottomrule
	\end{tabular} 
	\caption{Masked language modelling score examples.}
	\label{table:maskscore_examples}
\end{table}

\section{Further Experimental Results}
\label{sec:appendix:other_results}

See \autoref{table:all_mwoz_trained_results}, \autoref{table:number_of_phrases}, \autoref{table:trained_on_SGD_results} and \autoref{table:mwoz_test_result} for further experimental results. 


\begin{table*}[t]
	\centering
	\small
	\begin{tabular}{@{}lccccccc@{}}
		\toprule
		& \multicolumn{3}{c}{MultiWOZ} & \phantom{} & \multicolumn{3}{c}{SGD} \\
		\cmidrule{2-4} \cmidrule{6-8}
		Approach & F1-Score & Recall & Precision && F1-Score & Recall & Precision \\
		\midrule
		RoBERTa embeddings & 0.80 & 0.91 & 0.72 && 0.45 & 0.35 & 0.63 \\ 
		MLM scores & 0.38 & 0.83 & 0.25 && 0.34 & 0.34 & 0.35 \\
		Persistence image vectors & 0.53 & 0.87 & 0.38 && 0.47 & 0.46 & 0.48 \\ 
		Codensity & 0.42 & 0.76 & 0.29 && 0.37 & 0.34 & 0.42 \\ 
		Wasserstein norm & 0.37 & 0.65 & 0.26 && 0.42 & 0.40 & 0.44 \\
		TDA features together & 0.33 & 0.89 & 0.20 && 0.48 & 0.63 & 0.39 \\
		Union prediction & 0.28 & \textbf{0.96} & 0.17 && 0.48 & \textbf{0.74} & 0.36 \\
		\bottomrule
	\end{tabular} 
	\caption{
	    Results of all models trained on MultiWOZ and tested on MultiWOZ and SGD.
	    \label{table:all_mwoz_trained_results}
	}
\end{table*}


\begin{table*}[t]
	\centering
	\small
	\begin{tabular}{@{}lcc@{}}
		\toprule
		Approach & MultiWOZ & SGD  \\ 
		\midrule
		RoBERTa embeddings & 816 & 2757 \\
		MLM score & 2174 & 4933 \\
		Persistence image vectors & 1464 & 4775 \\
		Codensity & 1658 & 4054 \\
		Wasserstein norm & 1631 & 4536 \\
		TDA features & 2867 & 8189 \\
		Union prediction & 3712 & 10398 \\
		\bottomrule
	\end{tabular} 
	\caption{
	    Total number of terms tagged on MultiWOZ and SGD broken down per model trained on MultiWOZ.
	    For reference, there are 645 target terms in total in MultiWOZ and 5008 in SGD.
	    \label{table:number_of_phrases}
	}
\end{table*}

\begin{table*}
    \centering
    \small
	\begin{tabular}{@{}lccccccc@{}}
		\toprule
		& \multicolumn{3}{c}{MultiWOZ} & \phantom{} & \multicolumn{3}{c}{SGD} \\
		\cmidrule{2-4} \cmidrule{6-8}
		Approach & F1-Score & Recall & Precision && F1-Score & Recall & Precision \\
		\midrule
		RoBERTa embeddings & 0.65 & 0.92 & 0.50 && 0.83 & 0.88 & 0.78 \\ 
		MLM scores & 0.32 & 0.76 & 0.21 && 0.37 & 0.33 & 0.44 \\ 
		Persistence image vectors & 0.45 & 0.84 & 0.31 && 0.76 & 0.80 & 0.73 \\ 
		Codensity & 0.37 & 0.69 & 0.25 && 0.50 & 0.49 & 0.51 \\ 
		Wasserstein norm & 0.40 & 0.78 & 0.27 && 0.53 & 0.54 & 0.52 \\ 
		TDA features together & 0.30 & 0.92 & 0.18 && 0.64 & 0.88 & 0.50 \\
		Union prediction & 0.23 & \textbf{0.98} & 0.13 && 0.61 & \textbf{0.98} & 0.44 \\
		\bottomrule
	\end{tabular}
    \caption{
        Results of all models trained on SGD and tested on MultiWOZ and SGD.
        \label{table:trained_on_SGD_results}
    }
    
\end{table*}


\begin{table*}
    \centering
    \small
	\begin{tabular}{@{}lccc@{}}
		\toprule
		Approach & F1-Score & Recall & Precision \\
		\midrule
		RoBERTa embeddings & 0.87 & 0.91 & 0.84  \\
		MLM scores & 0.53 & 0.76 & 0.41 \\
		Persistence image vectors & 0.75 & 0.87 & 0.66  \\ 
		Codensity & 0.59 & 0.70 & 0.52  \\ 
		Wasserstein norm & 0.53 & 0.62 & 0.46 \\
		TDA features together & 0.57 & 0.92 & 0.41 \\
		Union prediction & 0.50 & \textbf{0.97} & 0.33 \\
		\bottomrule
	\end{tabular} 
	\caption{
	    Results of all models trained on MultiWOZ and tested on the MultiWOZ test set only.
	    \label{table:mwoz_test_result}
	}
\end{table*}

\section{Further Example Tags}
\label{sec:appendix:other_examples}

See \autoref{table:more_example_utterances_and_tags} for more utterances with the corresponding tags by the different models and \autoref{table:per_model_predictions_examples} for an analysis of which terms tagged by each model were already seen in MultiWOZ.

\begin{table*}[t]
	\centering
	\small
	\begin{tabular}{@{}p{3.6cm}|p{11.0cm}@{}}
		\toprule
		utterance & the curse of la llorona is a good one \\
		\midrule
		RoBERTa embeddings & \\
		MLM score & 
		\phantom{the curse of} \mybox[fill=blue!10]{la} \phantom{llorona is a} \mybox[fill=blue!10]{good} \mybox[fill=blue!10]{one} \\
		Persistence image vectors & 
		\phantom{the curse of} \mybox[fill=blue!10]{la llorona} \phantom{is a good one} \\
		Codensity & 
		\phantom{the curse of} \mybox[fill=blue!10]{la} \phantom{llorona is a good one}\\
		Wasserstein norm & \\ 
		\bottomrule
		
        \toprule
		utterance & i ' m bored . get me some tickets for an activity . \\
		\midrule
		RoBERTa embeddings & \\
		MLM score &  \\
		Persistence image vectors & 
		\phantom{i ' m bored . get me some tickets for an} \mybox[fill=blue!10]{activity} \phantom{.} \\
		Codensity & \\
		Wasserstein norm & \\ 
		\bottomrule
		
		\toprule
		utterance & what other therapists are there ? \\
		\midrule
		RoBERTa embeddings & \\
		MLM score &  \\
		Persistence image vectors &  \\
		Codensity & 
		\phantom{what other} \mybox[fill=blue!10]{therapists} \phantom{are there ?} \\
		Wasserstein norm & \\
		\bottomrule
		
		\toprule
		utterance & later on . for now i want to know the weather in there next wednesday . \\
		\midrule
		RoBERTa embeddings & 
		\phantom{later on . for now i want to know the weather in there next} \mybox[fill=blue!10]{wednesday} \phantom{.} \\
		MLM score & 
		\phantom{later on} \mybox[fill=blue!10]{.} \phantom{for now} \mybox[fill=blue!10]{i} \phantom{want to know the weather in there next} \mybox[fill=blue!10]{wednesday} \phantom{.} \\
		Persistence image vectors & 
		\phantom{later on . for now i want to know the weather in there next} \mybox[fill=blue!10]{wednesday} \phantom{.} \\
		Codensity & 
		\phantom{later on . for now i want to know the} \mybox[fill=blue!10]{weather} \phantom{in there next} \mybox[fill=blue!10]{wednesday} \phantom{.} \\
		Wasserstein norm & \\
		\bottomrule 
		
		\toprule
		utterance & do you know a place where i can get some food ? \\
		\midrule
		RoBERTa embeddings & 
		\phantom{do you know a} \mybox[fill=blue!10]{place} \phantom{where i can get some} \mybox[fill=blue!10]{food} \phantom{?} \\
		MLM score & 
		\phantom{do you know a place where i can get some} \mybox[fill=blue!10]{food} \phantom{?} \\
		Persistence image vectors & 
		\phantom{do you know a} \mybox[fill=blue!10]{place} \phantom{where i can get some} \mybox[fill=blue!10]{food} \phantom{?} \\
		Codensity & 
		\phantom{do you know a} \mybox[fill=blue!10]{place} \phantom{where i can get some} \mybox[fill=blue!10]{food} \phantom{?} \\
		Wasserstein norm & 
		\phantom{do you know a place where i can get some} \mybox[fill=blue!10]{food} \phantom{?} \\
		\bottomrule
		
		\toprule
		utterance & what time does the show begin ? \\
		\midrule
		RoBERTa embeddings & 
		\phantom{what} \mybox[fill=blue!10]{time} \phantom{does the show begin ?} \\
		MLM score & 
		\phantom{what time does the} \mybox[fill=blue!10]{show} \phantom{begin ?} \\
		Persistence image vectors & 
		\phantom{what} \mybox[fill=blue!10]{time} \phantom{does the show begin ?} \\
		Codensity & 
		\phantom{what} \mybox[fill=blue!10]{time} \phantom{does the show begin ?} \\
		Wasserstein norm & 
		\phantom{what} \mybox[fill=blue!10]{time} \phantom{does the show begin ?} \\
		\bottomrule
	\end{tabular} 
	\caption{
		More examples of tokenized utterances together with terms extracted by the different models.
		\label{table:more_example_utterances_and_tags}
	}
\end{table*}

\begin{table*}[t]
	\centering
	\small
	\begin{tabular}{@{}p{1.25cm}p{3.3cm}p{3.3cm}p{3.3cm}p{3.3cm}@{}}
		\toprule 
		Model & Seen in MultiWOZ & Only seen in SGD & False negatives & False positives  \\ 
		\midrule
		
		RoBERTa emb. & 
		Sushi Yoshizumi; Salesforce transit center; Jojo Restaurant \& Sushi Bar; bistro liaison; Eric's Restaurant; K\&L Bistro & Arizona vs. LA Dodgers; El Hombre; Arcadia; 795 El Camino Real; Owls vs. Tigers; Green Chile Kitchen;  & visit date; unapologetic; 134; The Motans; JT Leroy; Orchids Thai; 251 Llewellyn Avenue; 12221 San Pablo Avenue; Menara Kuala Lumpur; & Meriton; Rodeway Inn; Stewart; Embarcadero Center; Elysees; Shattuck; LAX; El; attractionin \\ \midrule
		
		MLM score & 
		350 Park Street; Doubletree by Hilton Hotel San Pedro - Port of Los Angeles; 24; Show Time; Up 2U Thai Eatery; 25; 381 South Van Ness Avenue; Broken English & Olly Murs; Bret Mckenzie; football game: USC vs Utah; stage door; 1012 Oak Grove Avenue; 'Mamma Mia; John R Saunderson; Alderwood Apartments & 630 Park Court; Unapologetic; visit date; The Motans; V's Barbershop Campbell; 101 South Front Street \#1; 134; Orchids Thai & humid then; others?; rad; wa; outdoor; alright, I; valley; spoke; webster; a song \\
		\midrule
		
		PI vectors & 
		Trademark Hotel; Dorsett City; London; Center Point Road O'Hare International Airport; Maya Palenque Restaurant; Casa Loma Hotel & Claude de Martino; Nero; Toronto FC vs Crew; Written in Sand; Emmylou Harris; Helen Patricia; Palo Alto Caltrain Station; Jack Carson & Shailesh Premi; Gorgasm; 157; Dad; destination city; serves alcohol 2556 Telegraph Avenue \#4; Glory Days; The Park Bistro \& Bar; Arcadia Sessions at The Presidio & Maggiano; XD; sexist scum; fir; red chillies; morning instead; capitol; Robin; !!! if so; free \\ 
		\midrule
		
		Codensity & 
		Tell me you love me; dentist name; The American Hotel Atlanta Downtown - A Doubletree by Hilton; Dim Sum Club; Le Apple Boutique Hotel KLCC; 555 Center Avenue & Hyatt Place New York/Midtown-South; colder weather; 'Little Mix; Commonwealth; 3630 Balboa Street; Newton Faulkner; directed by; How deep is your love & visit date; Unapologetic; 134; The Motans; V's Barbershop Campbell; 101 South Front Street \#1; Orchids Thai; 12221 San Pablo Avenue & and humid; vapour; 5:15; corect; names; flight leaving; collect; tiresome; Marriott \\ 
		\midrule
		
		Wasserstein norm & Wence's Restaurant; Miss me more; restaurant reservation; 1118 East Pike Street; El Charro Mexican Food \& Cantina; Murray Circle Restaurant & Broderick Roadhouse; Mets vs. Yankees; 226 Edelen Avenue; 1030; 162; Phillies vs. Cubs; 1110; Diamond Platnumz; '2664 Berryessa Road \#206; Oliveto & Anaheim Intermodal Center; Sangria; Vacation Inn Phoenix; 1776 First Street; After the Wedding; Mikey Day & loacation; enoteca; salone; balances; overseas; mars; help; Angeles and; 4:15; niles; titale; frmo; Oracle park \\
		\midrule
		
		TDA features together & 1300 University Drive \#6; The American Hotel Atlanta Downtown - A Doubletree by Hilton; Millennium Gloucester Hotel London Kensington & 4087 Peralta Boulevard; Power; Hyang Giri; Okkervil River; event location; 320; Jordan Smith; Caffe California; Ruth Bader Ginsburg; Neil Marshall; 171; 1599 Sanchez Street & Out of Love; Alderwood Apartments; has garage; 168; GP visit; Catamaran Resort Hotel and Spa; Dodgers vs. Diamondbacks; Showplace Icon Valley Fair; West Side Story & venu; being; replaced; parking; Okland; times; comments; pond; crowd; flick; 1,710; Blacow Road; Kathmandu \\ 
		\bottomrule
	\end{tabular} 
	\caption{Prediction examples of the different models on SGD (typos are reproduced as they appear in the data-set).}
	\label{table:per_model_predictions_examples}
\end{table*}

\end{document}